\documentclass{article}

\usepackage{iftex}
\ifPDFTeX
  \usepackage[utf8]{inputenc}
\fi
\usepackage{authblk}
\usepackage{main}
\usepackage{microtype}
\usepackage{times}
\usepackage{latexsym}

\usepackage{amsmath}
\usepackage{amssymb}
\usepackage{amsfonts}
\usepackage{mathtools}
\usepackage{bm}

\usepackage{graphicx}
\usepackage{fontawesome5}
\usepackage{subcaption}
\usepackage{float}
\usepackage{wrapfig}
\usepackage{booktabs}
\usepackage{array}
\usepackage{multirow}
\usepackage{makecell}
\usepackage{longtable}
\usepackage{tabularx}
\usepackage{ragged2e}
\usepackage{placeins}

\usepackage{xcolor}
\definecolor{mydarkblue}{rgb}{0,0.08,0.45}
\usepackage{enumitem}
\usepackage{geometry}
\usepackage{fancyhdr}
\usepackage{setspace}

\usepackage{listings}

\usepackage{natbib}
\usepackage{url}
\usepackage[
  colorlinks=true,
  linkcolor=mydarkblue,
  citecolor=mydarkblue,
  filecolor=mydarkblue,
  urlcolor=mydarkblue
]{hyperref}

\newenvironment{itemize*}
  {\begin{itemize}[leftmargin=*,nosep]}
  {\end{itemize}}
\newenvironment{enumerate*}
  {\begin{enumerate}[leftmargin=*,nosep]}
  {\end{enumerate}}

\title{\textsc{RiboSpan}: A Long-Context RNA Foundation Model\\for Versatile RNA Modeling}

\author[1,2,4,5]{Ziyuan Wang\textsuperscript{\textdagger}}
\author[1,3,5]{Bohao Tang\textsuperscript{\textdagger}}
\author[1,3,5]{Fei Zhang}
\author[2,4]{Shuo Han\textsuperscript{*}}
\author[1,3,5]{Pengfei Liu\textsuperscript{*}}
\affil[1]{Shanghai Innovation Institute}
\affil[2]{Center for Excellence in Molecular Cell Science, CAS}
\affil[3]{Shanghai Jiao Tong University}
\affil[4]{University of Chinese Academy of Sciences}
\affil[5]{Generative Artificial Intelligence Research Lab}

\date{}

\fancypagestyle{reporttitle}{
  \fancyhf{}
  \lhead{%
    \raisebox{-0.35cm}{\includegraphics[height=0.96cm]{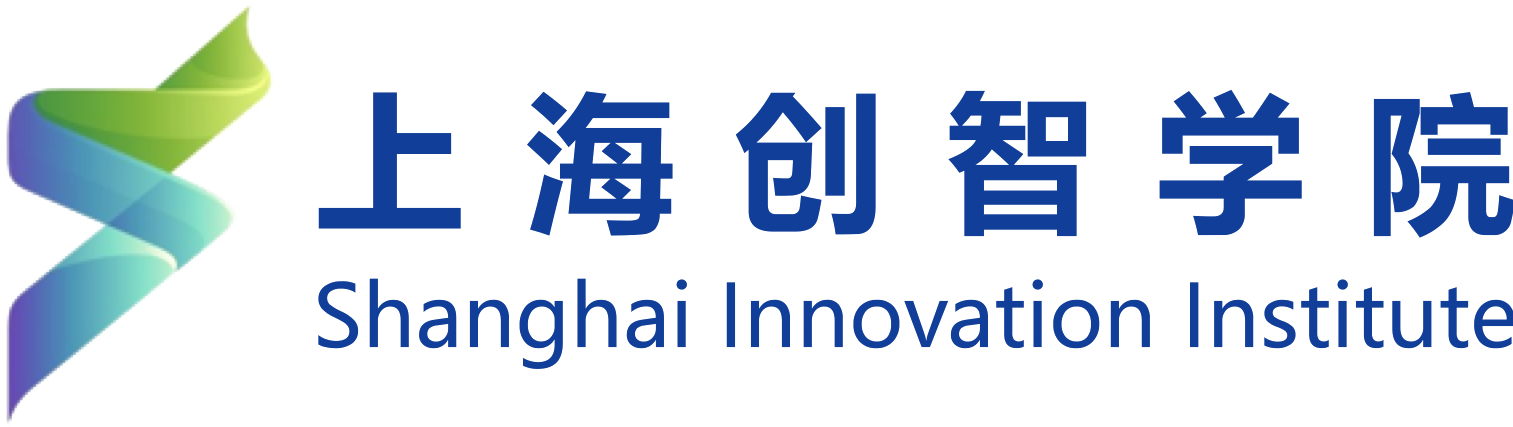}}%
    \hspace{0.22cm}%
    \raisebox{-0.45cm}{\includegraphics[height=1.10cm]{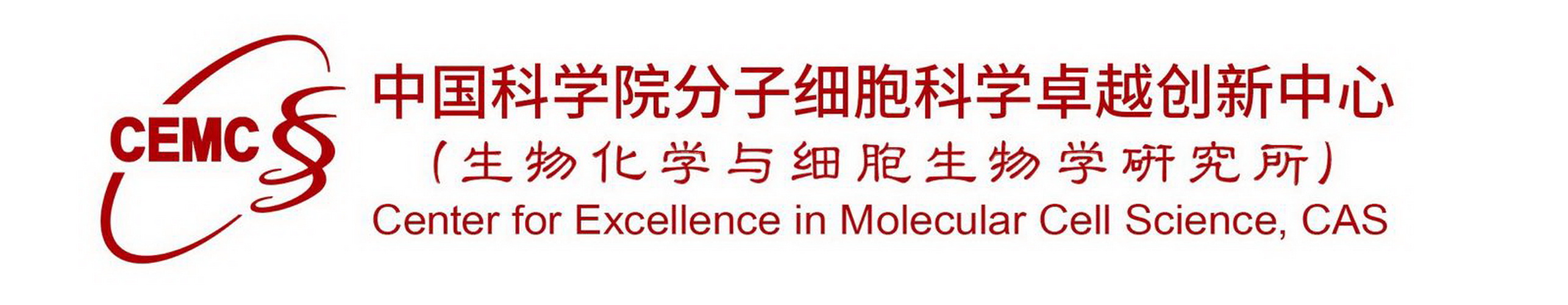}}%
    \hspace{0.10cm}%
    \raisebox{-0.38cm}{\includegraphics[height=1.01cm]{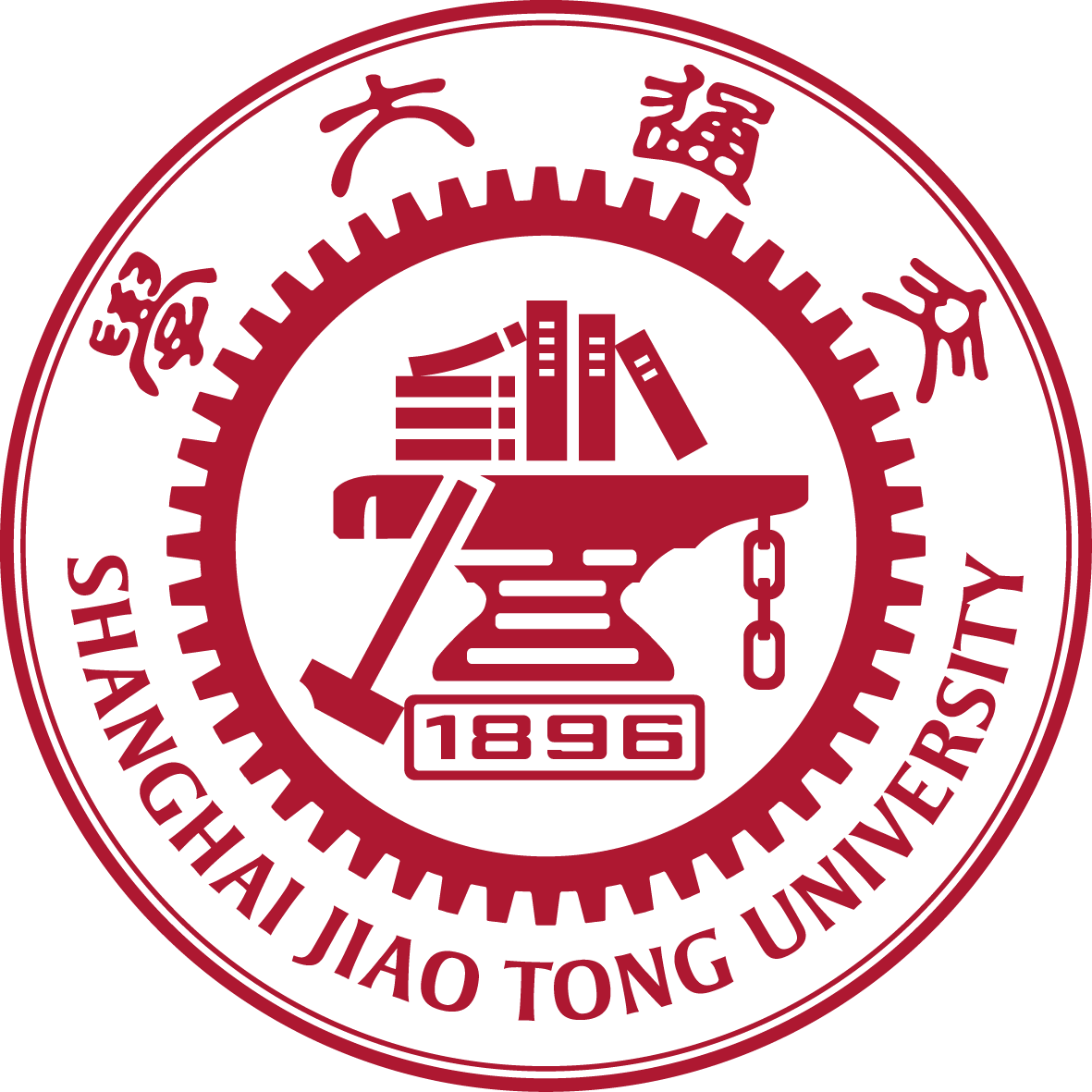}}%
    \hspace{0.50cm}%
    \raisebox{-0.38cm}{\includegraphics[height=1.01cm]{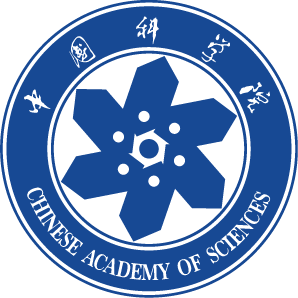}}%
  }
  \rhead{%
    \raisebox{-0.28cm}{\includegraphics[height=0.86cm]{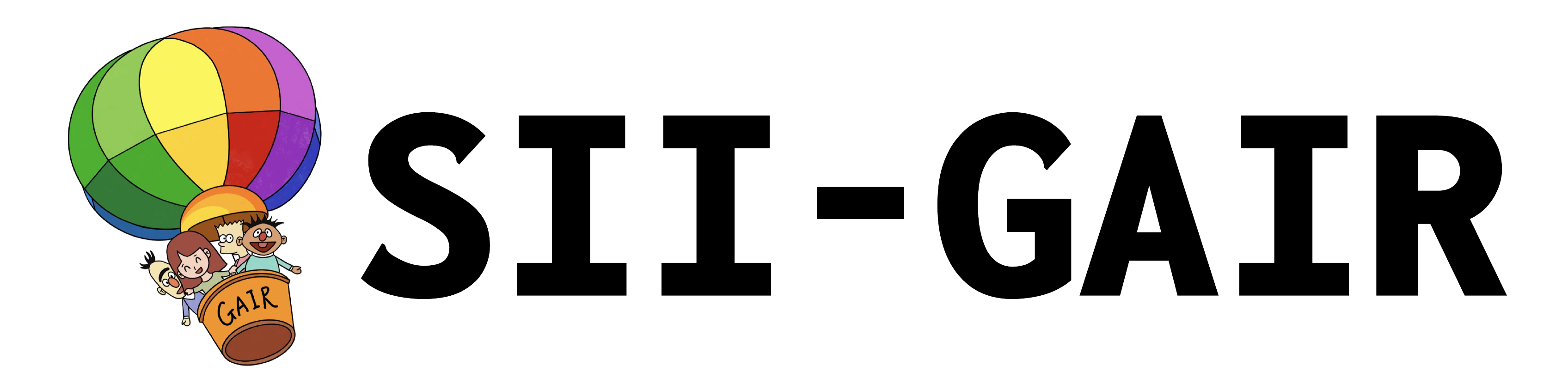}}%
  }
  \cfoot{\thepage}
  
}

\usepackage{adjustbox}
\begin{document}

\maketitle
\thispagestyle{reporttitle}

{\centering
\href{https://github.com/GAIR-NLP/RIBOSPAN-FM}
{\textcolor{black}{\faGithub}\ GitHub}
\quad
\href{https://huggingface.co/SII-GAIR-NLP/RIBOSPAN-FM}
{\raisebox{-.15em}{\includegraphics[height=1em]{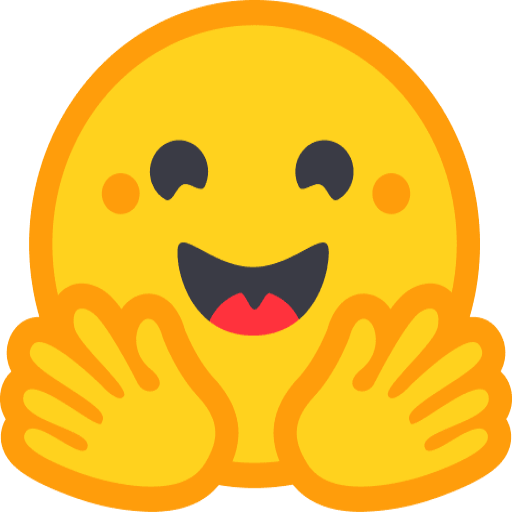}}\ Hugging Face}
\par}

\vspace{0.4em}

\begin{center}
  \small
  \textsuperscript{\textdagger}Equal contribution.
  \quad
  \textsuperscript{*}Co-corresponding authors.
\end{center}



\vspace{1em}


\begin{abstract}
Full-length RNAs, particularly messenger RNAs, often exceed the context lengths used to pretrain existing RNA foundation models, limiting complete-transcript modeling at single-nucleotide resolution. We present \textsc{RiboSpan}, a 1.61-billion-parameter bidirectional RNA foundation model natively pretrained with context lengths up to 10,240~nt, combining dense bidirectional self-attention with single-nucleotide tokenization. On a controlled long-context benchmark, native long-context pretraining maintains well-organized, context-specific representations across full transcripts, an advantage that inference-time extension of short-context models does not recover. \textsc{RiboSpan} learns state-of-the-art frozen RNA representations, and is the strongest encoder-only RNA foundation model in full-transcript property prediction and zero-shot mutation-fitness modeling. Without structural supervision during pretraining, its frozen backbone outperforms a fully fine-tuned structure-guided model on secondary structure prediction. Building on the same backbone, we develop a conditional discrete-diffusion framework for full-length mRNA generation and redesign. Together, \textsc{RiboSpan} establishes a powerful long-context foundation for transferable RNA representation learning, biological prediction, and full-transcript mRNA design.
\medskip
\end{abstract}

\section{Introduction}
\label{sec:introduction}

Messenger RNA (mRNA) is a biomolecule with a modularly annotated architecture but highly coupled functions across its constituent regions. The 5$^{\prime}$ untranslated region (5$^{\prime}$ UTR), coding sequence (CDS), and 3$^{\prime}$ untranslated region (3$^{\prime}$ UTR) jointly influence multiple aspects of translation initiation and elongation, RNA folding, molecular stability, subcellular localization, and interactions with RNA-binding proteins and other cellular components. Consequently, a sequence alteration within one region may affect molecular phenotypes typically attributed to another region through structural rearrangements or changes in the broader regulatory context. Such coupling is particularly important for therapeutic and synthetic mRNAs, for which a candidate sequence must not only encode the correct protein but also satisfy multiple constraints related to translational efficiency, stability, and manufacturability~\citep{mauger2019structure,leppek2022combinatorial,metkar2024tailormade}.

Existing computational methods generally make this high-dimensional design problem tractable by constraining the sequence region or optimization objective. Substantial progress has been made in 5$^{\prime}$ UTR activity prediction and optimization~\citep{sample2019utr}, joint optimization of synonymous codon usage and RNA structure~\citep{zhang2023lineardesign}, and generative mRNA design~\citep{zhang2025gemorna,patel2026mrnautilus}. These approaches nevertheless address different parts of the broader modeling problem. Region-specific methods optimize selected regulatory elements or functional readouts, while synonymous-CDS optimization preserves the encoded protein and searches within a restricted sequence space. GEMORNA extends generative design across CDS and UTRs but models these transcript regions separately, whereas mRNAutilus performs joint full-transcript generation through a masked discrete-diffusion objective oriented toward sequence denoising and property-guided optimization rather than transferable representation learning. Together, these advances substantially expand the scope of computational mRNA design, while leaving complete-transcript representation learning less systematically explored.

RNA foundation models offer a promising route toward this broader objective. Bidirectional encoders such as RNA-FM~\citep{chen2022rnafm}, RiNALMo~\citep{penic2025rinalmo}, and AIDO.RNA~\citep{zou2024aido} have shown that large-scale masked nucleotide modeling can learn transferable representations associated with RNA structure and function. Bidirectional self-attention~\citep{vaswani2017attention,devlin2019bert} allows each nucleotide to integrate both upstream and downstream context, making this architecture naturally suited to full-sequence understanding tasks such as structure prediction, functional prediction, and per-nucleotide representation learning. Recent encoders extend this paradigm in several directions: RiboGRAM~\citep{wang2026ribogram} augments self-attention with multi-scale convolutional encoding, LIMBER~\citep{wang2026limber} interleaves attention with state-space layers, and structRFM~\citep{zhu2026structrfm} and RNAverse~\citep{ma2026rnaverse} incorporate structural information into pretraining through structure-guided masking and experimental structure-probing profiles, respectively. However, these encoders are typically pretrained with context lengths of at most approximately 1,024 tokens; sequences exceeding this length must therefore be truncated or cropped, making it difficult to preserve the complete context of many mature mRNAs.

The need to model complete transcripts has driven the development of long-sequence RNA foundation models along two main directions: generation and representation. The first comprises decoder-only models exemplified by EVA~\citep{huang2026eva}. The causal attention mechanism is well suited to autoregressive generation and has extended sequence scoring, \textit{de novo} generation, and targeted regional redesign to the transcript scale. Under causal attention, however, each position can access only its upstream sequence, preventing per-nucleotide representations from simultaneously integrating the complete context on both sides of a given position. RNA structure formation, molecular binding, and regulatory activity often involve relationships among both local and distal sequence elements. Unidirectional information flow is therefore not fully aligned with the representation of complete RNA structure and function. For this reason, many representative foundation models for sequence understanding and downstream prediction across proteins, DNA, and RNA adopt bidirectional masked language modeling to learn contextual representations~\citep{hayes2025esm3,chen2022rnafm,dallatorre2025nucleotide}.

The second direction comprises encoder-only models designed for bidirectional representation of long RNAs. These models preserve the joint use of upstream and downstream information while expanding sequence coverage by reducing computational complexity or input tokens. HydraRNA primarily uses bidirectional state-space modules while retaining multi-head attention in selected layers; RNAret adopts a linear-complexity bidirectional retention mechanism; and BiRNA-BERT applies byte-pair encoding to long inputs, compressing multiple consecutive nucleotides into fewer tokens~\citep{li2025hydrarna,shen2026rnaret,tahmid2025birnabert}. These designs substantially reduce the computational cost of long-sequence modeling but introduce corresponding representational trade-offs: efficient sequence-mixing architectures do not retain dense all-to-all nucleotide interactions in every layer, whereas token compression sacrifices fixed single-nucleotide resolution for long inputs.

Together, these two lines of work have advanced transcript-scale generation and bidirectional representation of long RNAs. Nevertheless, a key capability gap remains in complete mRNA modeling: existing models have not yet simultaneously achieved single-nucleotide tokenization, dense bidirectional self-attention throughout all layers, and billion-scale model capacity at a context length representative of full-length transcripts~\citep{chen2022rnafm,zou2024aido,penic2025rinalmo,li2025hydrarna,tahmid2025birnabert,shen2026rnaret,huang2026eva,wang2026ribogram,wang2026limber,zhu2026structrfm,ma2026rnaverse}. For complete transcripts, these three properties jointly preserve per-nucleotide positional resolution, support comprehensive integration of upstream, downstream, and distal sequence information, and place the 5$^{\prime}$ UTR, CDS, 3$^{\prime}$ UTR, and their cross-region dependencies within a unified high-resolution representation space. RoPE-based context-extension methods such as Position Interpolation and YaRN can enlarge the usable positional range of pretrained models~\citep{su2024roformer,chen2023position,peng2024yarn}. However, positional extension alone does not expose the backbone during pretraining to sequence interactions at full-transcript lengths. It therefore cannot replace native long-context pretraining when the goal is to learn representations from complete RNA sequences.

\medskip

To address this capability gap, our main contributions are as follows:

\begin{itemize}[leftmargin=*]
  \item First, we trained \textsc{RiboSpan}, the first billion-scale dense bidirectional Transformer RNA foundation model with single-nucleotide tokenization and native pretraining up to 10,240 nt. It learns transferable representations across diverse RNA types while extending high-resolution RNA modeling to full-length transcripts.

  \item Second, we introduced the first benchmark for systematically evaluating long-context representations in RNA foundation models. It evaluates whether representations remain contextually organized as sequence length extends beyond the pretrained context and, through an attention-restriction ablation, identifies the source of the failure of short-context models and what native long-context pretraining contributes beyond it.

  \item Third, we developed the first full-stack mRNA design framework built on a long-context RNA foundation model, enabling joint full-length generation and transcript-wide redesign. Using \textsc{RiboSpan} as the generative backbone, the framework jointly models the 5$^\prime$ UTR, CDS, and 3$^\prime$ UTR under transcript context, with multidimensional conditioning and synonymous-codon diffusion for cross-region and protein-preserving optimization.
\end{itemize}


\section{Native Long-Context Pretraining}
\label{sec:model}

\textsc{RiboSpan} uses a 1.61B-parameter bidirectional Transformer encoder with single-nucleotide tokenization. The pretraining corpus combines diverse RNA sequences from RNAcentral with annotated protein-coding transcripts from Ensembl. Models are pretrained with a native context length of 10,240 tokens, and an initial 15\% masking stage is followed by continued pretraining with 40\% masking, providing long-context foundation models for nucleotide-level RNA representation learning and generative modeling.

\subsection{Model Architecture and Tokenization}

\begin{figure}[H]
  \centering
  \includegraphics[width=0.8\linewidth]{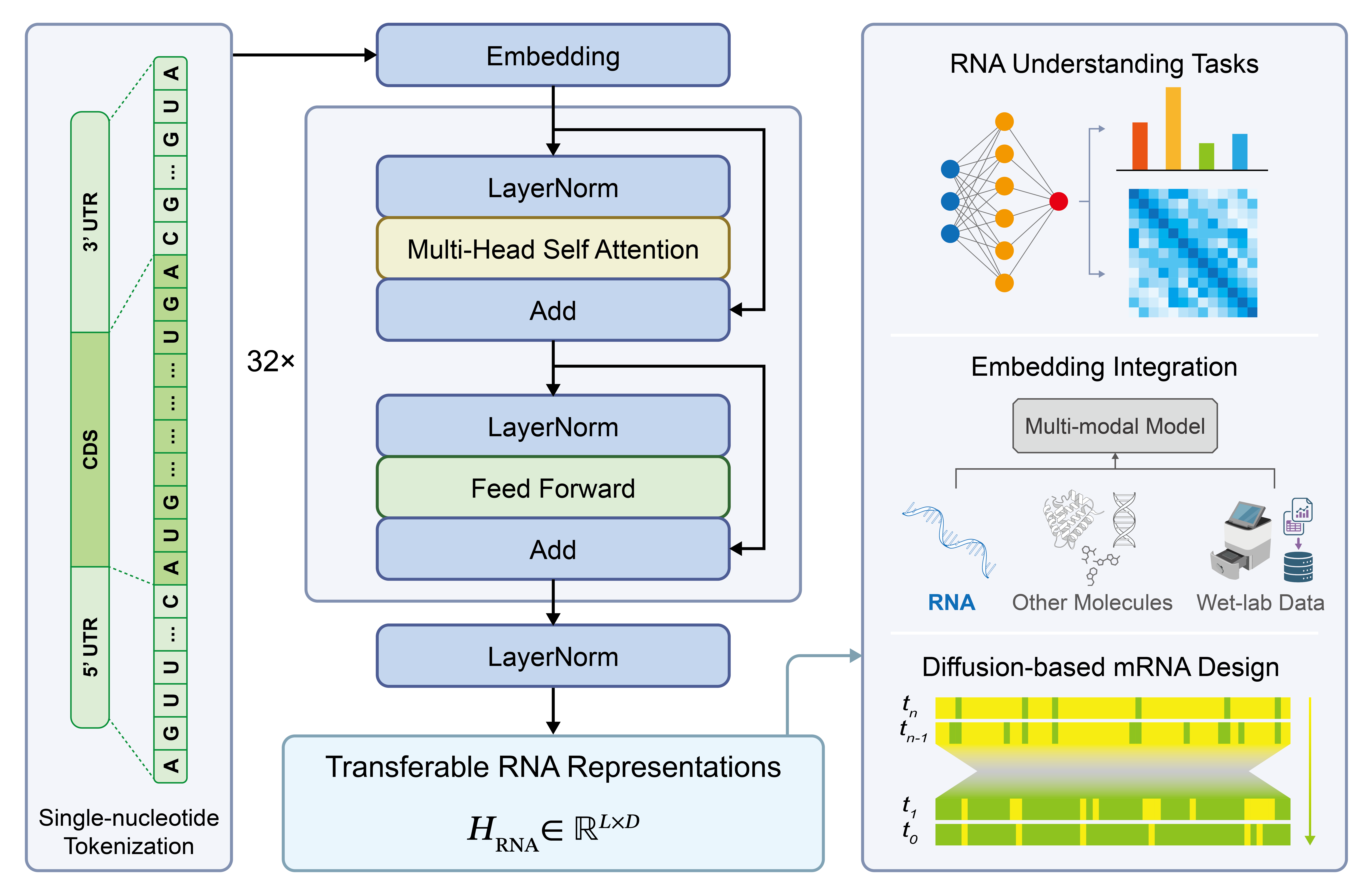}
  \caption{Architecture and downstream use of \textsc{RiboSpan}. Single-nucleotide RNA sequences are encoded by 32 bidirectional Transformer layers into transferable nucleotide-level representations for downstream RNA modeling.}
  \label{fig:model-architecture}
\end{figure}

\textsc{RiboSpan} uses a 32-layer pre-norm Transformer encoder with a dimension of 2,048 (Figure~\ref{fig:model-architecture}). Each transformer block consists of bidirectional multi-head self-attention followed by a SwiGLU feed-forward network. RoPE~\citep{su2024roformer} is applied over the full attention-head dimension. Table~\ref{tab:model-config} summarizes the configuration.

\begin{table}[H]
  \centering
  \caption{Backbone configuration of \textsc{RiboSpan}.}
  \label{tab:model-config}
  \small
  \begin{tabular}{ll@{\hspace{1.3cm}}ll}
    \toprule
    Setting & Configuration & Setting & Configuration \\
    \midrule
    Transformer Layers & 32 & Model Dimension & 2,048 \\
    FFN Intermediate size & 5,440 & Attention Heads & 32 \\
    Activation & SwiGLU & Normalization & LayerNorm \\
    Position Encoding & RoPE (rotary dim = 64) & Head Dimension & 64 \\
    Vocabulary Size & 16 & Token Unit & nucleotide \\
    Native Context Length & 10,240 tokens & Parameters & 1.61B \\
    \bottomrule
  \end{tabular}
\end{table}

\textsc{RiboSpan} uses single-nucleotide tokenization, with each nucleotide occupying one token position. Vocabulary construction, sequence normalization, and boundary-token handling are detailed in Appendix~\ref{app:training-details}.

\subsection{Pretraining Corpus}

We construct the pretraining corpus from RNAcentral v26.0~\citep{rnacentral2026}, Ensembl release 115, and Ensembl Genomes release 62 \citep{dyer2025ensembl}. After source-specific filtering, normalization, and exact deduplication, the final training corpus contains 67.6M RNA sequences and 85.7B nucleotide tokens in total.

\begin{figure}[H]
  \centering
  \includegraphics[width=\linewidth]{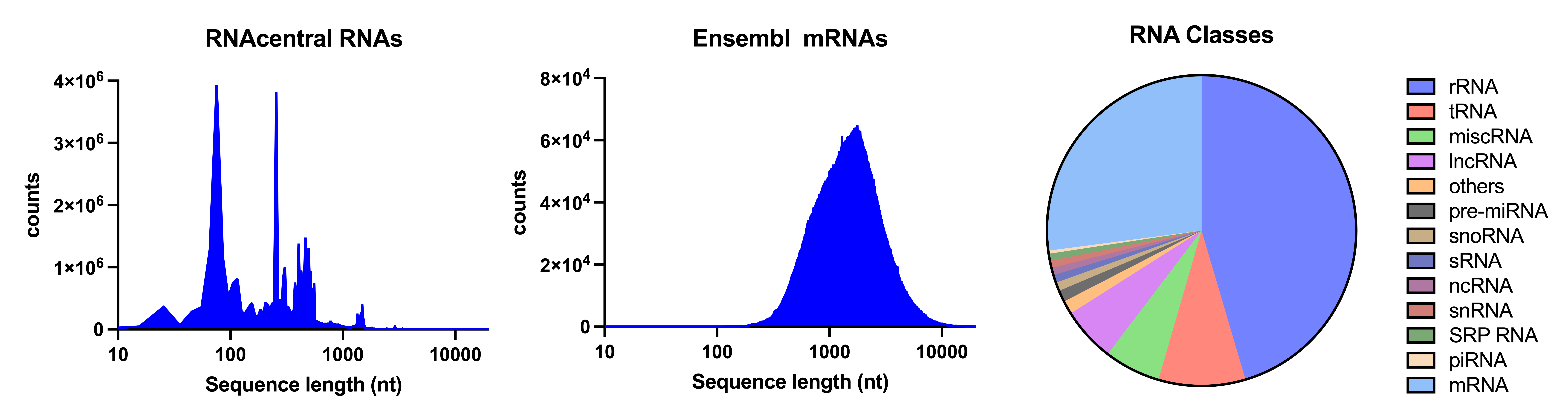}
  \caption{Sequence-length distributions and RNA-class composition of the pretraining corpus.}
  \label{fig:corpus-composition}
\end{figure}

RNAcentral provides broad coverage across RNA classes, while Ensembl datasets contribute quality-controlled protein-coding transcripts with complete CDS and UTR annotations across vertebrates, plants, fungi, metazoans, and protists. Source-specific filtering, normalization, deduplication, and held-out set construction are described in Appendix~\ref{app:data-curation}. The resulting sequence-length and RNA-class distributions are shown in Figure~\ref{fig:corpus-composition}.

The validation and test sets were constructed using class-specific sampling quotas to provide broad representation across RNA classes while reducing the influence of the highly imbalanced training distribution. For protein-coding transcripts, the sampling was further stratified by species. The RNA-class composition of the training corpus and held-out sets is summarized in Table~\ref{tab:data-composition}.

\begin{table}[H]
  \centering
  \caption{RNA-class composition of the pretraining corpus and held-out sets.}
  \label{tab:data-composition}
  \small
  \begin{tabular}{lrrr}
    \toprule
    RNA Class & Training & Validation & Test \\
    \midrule
    rRNA       & 30,748,160 & 10,000 & 10,000 \\
    mRNA       & 18,316,110 & 10,000 & 10,000 \\
    tRNA       &  6,135,163 & 10,000 & 10,000 \\
    miscRNA  &  3,966,721 & 10,000 & 10,000 \\
    lncRNA     &  3,818,804 & 10,000 & 10,000 \\
    others     &    940,161 &  5,000 &  5,000 \\
    pre-miRNA &    804,307 &  5,000 &  5,000 \\
    snoRNA     &    598,283 &  5,000 &  5,000 \\
    sRNA       &    541,829 &  5,000 &  5,000 \\
    ncRNA      &    510,635 &  5,000 &  5,000 \\
    snRNA      &    493,584 &  5,000 &  5,000 \\
    SRP-RNA   &    483,594 &  5,000 &  5,000 \\
    piRNA      &    209,734 &  5,000 &  5,000 \\
    \midrule
    Total      & 67,567,085 & 90,000 & 90,000 \\
    \bottomrule
  \end{tabular}
\end{table}

\subsection{Long-Context Pretraining Schedule}

\textsc{RiboSpan} is pretrained with masked language modeling (MLM) at a native context length of 10,240 tokens. We first pretrain the model with a 15\% masking rate and then continue pretraining from the resulting checkpoint with a higher 40\% masking rate for reconstruction-oriented adaptation.

To evaluate the effect of native context length, we additionally train 1,024-token baseline models using the same corpus, architecture, and masking schedule. Four model variants are summarized in Table~\ref{tab:model-variants}.

\begin{table}[H]
  \centering
  \caption{\textsc{RiboSpan} pretraining variants.}
  \label{tab:model-variants}
  \small
  \begin{tabular}{lccl}
    \toprule
    Model Variant & Native Context Length & Masking Rate & Training Stage \\
    \midrule
    \textsc{RiboSpan}-10K-15 & 10,240 & 15\% & Pretraining \\
    \textsc{RiboSpan}-10K-40 & 10,240 & 40\% & Continued pretraining from 10K-15 \\
    \textsc{RiboSpan}-1K-15  & 1,024  & 15\% & Pretraining \\
    \textsc{RiboSpan}-1K-40  & 1,024  & 40\% & Continued pretraining from 1K-15 \\
    \bottomrule
  \end{tabular}
\end{table}

Pretraining is implemented with Megatron-LM~\citep{shoeybi2019megatron}, with customization to its BERT pretraining pipeline for variable-length sequence packing. Multiple RNAs are packed into a common training sequence while retaining independent sequence boundaries; attention is restricted within each RNA, and positional indices are reset at sequence boundaries, preventing information leakage across packed sequences. Padding tokens are excluded from the training objective. Details on masking, packing, and optimization are provided in Appendix~\ref{app:training-details}.

\FloatBarrier


\section{Experiments}
\label{sec:experiments}

\subsection{mRNA Reconstruction Evaluation}
\label{sec:reconstruction}

We evaluate nucleotide reconstruction on the mRNA subset of the pretraining test split with maximum input lengths of 1,024 and 10,240 tokens under masking rates of 15\% and 40\%. The four \textsc{RiboSpan} variants defined in Table~\ref{tab:model-variants} are evaluated together with AIDO.RNA-CDS, a coding-sequence-adapted variant of the AIDO.RNA foundation model~\citep{zou2024aido}, as a reference. At 10,240 tokens, AIDO.RNA-CDS and the \textsc{RiboSpan}-1K variants are evaluated through direct RoPE extrapolation, whereas the \textsc{RiboSpan}-10K variants remain within their native context. The 15\% and 40\% masking rates represent standard and heavy corruption settings, respectively.

\paragraph{Masked-Language-Modeling Loss.}
For an original sequence $x=(x_1,\ldots,x_L)$ and its corrupted input $\widetilde{x}$, let $M$ denote the nucleotide positions selected for prediction. The masked language modeling loss is
\begin{equation}
\mathcal{L}_{\mathrm{MLM}}
=-\frac{1}{|M|}\sum_{i\in M}\log p_{\theta}\!\left(x_i\mid\widetilde{x}\right),
\label{eq:reconstruction-mlm}
\end{equation}
where $p_{\theta}(x_i\mid\widetilde{x})$ is the probability assigned to the original nucleotide at position $i$, and $M$ contains 15\% or 40\% of the valid nucleotide positions according to the evaluation setting.

\begin{table}[H]
  \centering
  \caption{Masked language modeling loss on the mRNA subset of the pretraining test split. Bold and underlined values in result tables indicate the best and second-best performance, respectively.}
  \label{tab:mlm-loss-full}
  \footnotesize
  \begin{tabular}{lcccc}
    \toprule
    & \multicolumn{2}{c}{15\% Masking} & \multicolumn{2}{c}{40\% Masking} \\
    \cmidrule(lr){2-3}\cmidrule(lr){4-5}
    Model & 1,024 tokens & 10,240 tokens & 1,024 tokens & 10,240 tokens \\
    \midrule
    AIDO.RNA-CDS~\citep{zou2024aido} & 1.08489 & 1.15072 & 1.13499 & 1.19024 \\
    \textsc{RiboSpan}-1K-15 & \textbf{0.67434} & 0.98153 & 0.93860 & 1.27385 \\
    \textsc{RiboSpan}-1K-40 & \underline{0.68036} & 1.02138 & \textbf{0.77438} & 1.06730 \\
    \textsc{RiboSpan}-10K-15 & 0.76147 & \textbf{0.72417} & 0.99496 & \underline{0.91122} \\
    \textsc{RiboSpan}-10K-40 & 0.75660 & \underline{0.72519} & \underline{0.88394} & \textbf{0.80033} \\
    \bottomrule
  \end{tabular}
\end{table}

Reconstruction loss reveals a clear context-length effect. Direct extrapolation of short-context models degrades substantially at 10,240 tokens, whereas the native 10K variants retain strong reconstruction performance. The 40\% masking continuation further improves recovery under heavy corruption with little change under 15\% masking.

\paragraph{Global Reconstruction Accuracy.}
MLM loss evaluates reconstruction only at masked positions and therefore does not reveal whether contextual encoding preserves the identity of nucleotides that remain visible in the input. We additionally report sequence-wide argmax recovery over all valid nucleotide positions as a diagnostic of representation fidelity, testing whether the model can reconstruct masked nucleotides while retaining accurate nucleotide identity at observed positions. Global reconstruction accuracy is defined as:
\begin{equation}
\mathit{Acc}_{\mathrm{global}}
=
\frac{1}{L}
\sum_{i=1}^{L}
\mathbf{1}\!\left[\hat{x}_i=x_i\right],
\qquad
\hat{x}_i
=
\underset{x\in\mathrm{Vocab}}{\arg\max}\;
p_{\theta}\!\left(x\mid\widetilde{x},i\right).
\label{eq:global-reconstruction}
\end{equation}

\begin{table}[H]
  \centering
  \caption{Global nucleotide reconstruction accuracy on the mRNA subset of the pretraining test split. Bold and underlined values in result tables indicate the best and second-best performance, respectively.}
  \label{tab:mlm-accuracy-full}
  \footnotesize
  \begin{tabular}{lcccc}
    \toprule
    & \multicolumn{2}{c}{15\% Masking} & \multicolumn{2}{c}{40\% Masking} \\
    \cmidrule(lr){2-3}\cmidrule(lr){4-5}
    Model & 1,024 tokens & 10,240 tokens & 1,024 tokens & 10,240 tokens \\
    \midrule
    AIDO.RNA-CDS~\citep{zou2024aido} & 0.91953 & 0.91576 & 0.79203 & 0.77791 \\
    \textsc{RiboSpan}-1K-15 & \textbf{0.94973} & 0.93083 & 0.83343 & 0.77168 \\
    \textsc{RiboSpan}-1K-40 & \underline{0.94602} & 0.92589 & \textbf{0.86076} & 0.80264 \\
    \textsc{RiboSpan}-10K-15 & 0.94349 & \textbf{0.94763} & 0.82015 & \underline{0.83769} \\
    \textsc{RiboSpan}-10K-40 & 0.94163 & \underline{0.94517} & \underline{0.84094} & \textbf{0.85887} \\
    \bottomrule
  \end{tabular}
\end{table}

Global reconstruction accuracy confirms the same pattern. The native 10K variants retain high recovery accuracy at 10,240 tokens, while the reconstruction-oriented 40\% continuation improves recovery under heavy masking at both context lengths. In particular, \textsc{RiboSpan}-10K-40 combines the long-context capability established by native 10K pretraining with substantially stronger robustness to severe corruption, while remaining closely matched to 10K-15 under 15\% masking at both input lengths.

\FloatBarrier

\subsection{Long-Context Representation Benchmark}
\label{sec:benchmark}

To systematically evaluate how RNA foundation models represent long sequences, we construct a long-context benchmark based on complete mRNA transcripts. The benchmark characterizes whether the representations remain contextually organized as the input length increases, through contextual responsiveness, region-specific representation organization, and the spatial extent of perturbation-induced representation changes. These representation measures are complemented by masked-nucleotide accuracy on the same transcripts, and by an attention-restriction ablation that identifies the source of the failures observed beyond the pretrained context and what native long-context pretraining contributes beyond it.

Specifically, we examine how models respond to a localized composition-preserving sequence rearrangement, how distinctly they represent the modified and surrounding sequence contexts, and whether the resulting representation changes remain localized or propagate into distant unchanged regions. We further measure how accurately each model recovers masked nucleotides of the native transcripts, as an indication of how well the available context is used for prediction. Together, these measurements enable direct comparison among short-context models, inference-time position-extended models, efficient long-sequence architectures, and models natively pretrained with long contexts across increasing sequence lengths.

\subsubsection{Benchmark Design}
\label{sec:benchmark-design}

The benchmark comprises complete mRNAs spanning five length groups of 1,024, 2,048, 4,096, 8,192, and 10,240 nt, with 10 transcripts per group. For a transcript of length $\mathrm{L}$, a centered interval of width $W=\operatorname{round}(\mathrm{L}/32)$ is used to construct a native-structured pair. The structured sequence reorders nucleotides only within this interval while preserving its nucleotide composition and leaving all positions outside the interval unchanged.

We evaluate the four \textsc{RiboSpan} variants defined in Table~\ref{tab:model-variants} together with four external references: HydraRNA~\citep{li2025hydrarna}, AIDO.RNA-CDS~\citep{zou2024aido}, RiboGRAM~\citep{wang2026ribogram}, and RNAverse~\citep{ma2026rnaverse}. HydraRNA uses a hybrid state-space/attention architecture and is evaluated directly at each requested sequence length. AIDO.RNA-CDS, RiboGRAM, and RNAverse are short-context models with rotary position embeddings and a pretrained context of 1,024 tokens; RiboGRAM additionally combines self-attention with parallel multi-scale convolutional branches. Like the two \textsc{RiboSpan}-1K variants, they are evaluated beyond their pretrained range by direct RoPE extrapolation. For AIDO.RNA-CDS and the two \textsc{RiboSpan}-1K variants, we additionally evaluate dynamic YaRN positional scaling~\citep{peng2024yarn} at inference using the same model weights without additional training. Model-specific long-sequence evaluation settings are detailed in Appendix~\ref{app:position-extension}.

\subsubsection{Evaluation Measures}

\paragraph{Context Separation.}
Context Separation measures how strongly the same nucleotide is represented differently between the intervention interval and the surrounding background. To isolate contextual effects from nucleotide identity, the score is computed separately for each nucleotide type.

For each sequence in a pair, let $\mathcal{I}$ and $\mathcal{B}$ denote the sets of positions in the intervention interval and surrounding background, respectively, with the 8-nt buffer excluded from $\mathcal{B}$. Let $\mathcal{V}=\{\mathrm{A},\mathrm{C},\mathrm{G},\mathrm{T}\}$ denote the nucleotide alphabet. For each $b\in\mathcal{V}$, define
$\mathcal{I}_b=\{i\in\mathcal{I}:x_i=b\}$ and
$\mathcal{B}_b=\{i\in\mathcal{B}:x_i=b\}$,
where $\mathbf{h}_i^{(\ell)}$ denotes the representation of position $i$ at layer $\ell$.

At each layer, we average the cosine similarity over three types of position pairs: pairs within the intervention interval, pairs within the background, and pairs spanning the two regions:
\begin{align}
C_{\mathcal{I}}^{(b,\ell)}
&=
\mathbb{E}_{\substack{i,j\in\mathcal{I}_b\\i\neq j}}
\left[
\cos\left(\mathbf{h}_i^{(\ell)},\mathbf{h}_j^{(\ell)}\right)
\right],\\
C_{\mathcal{B}}^{(b,\ell)}
&=
\mathbb{E}_{\substack{i,j\in\mathcal{B}_b\\i\neq j}}
\left[
\cos\left(\mathbf{h}_i^{(\ell)},\mathbf{h}_j^{(\ell)}\right)
\right],\\
C_{\mathcal{I}\mathcal{B}}^{(b,\ell)}
&=
\mathbb{E}_{\substack{i\in\mathcal{I}_b\\j\in\mathcal{B}_b}}
\left[
\cos\left(\mathbf{h}_i^{(\ell)},\mathbf{h}_j^{(\ell)}\right)
\right].
\end{align}

The two within-region similarities are combined into a pair-count-weighted same-region baseline:
\begin{equation}
C_{\mathrm{same}}^{(b,\ell)}
=
\frac{
n_{\mathcal{I}}^{(b)}C_{\mathcal{I}}^{(b,\ell)}
+
n_{\mathcal{B}}^{(b)}C_{\mathcal{B}}^{(b,\ell)}
}{
n_{\mathcal{I}}^{(b)}+n_{\mathcal{B}}^{(b)}
},
\end{equation}
where $n_{\mathcal{I}}^{(b)}$ and $n_{\mathcal{B}}^{(b)}$ denote the corresponding numbers of within-region pairs.

Context Separation for nucleotide $b$ is then defined as the difference between the same-region baseline and the corresponding cross-region similarity:
\begin{equation}
\mathit{CS}^{(b,\ell)}
=
C_{\mathrm{same}}^{(b,\ell)}
-
C_{\mathcal{I}\mathcal{B}}^{(b,\ell)}.
\end{equation}

At the final layer, Additional Context Separation ($\Delta\mathit{CS}$) measures the increase in regional separation induced by the composition-preserving rearrangement:
\begin{equation}
\Delta\mathit{CS}
=
\frac{1}{|\mathcal{V}|}
\sum_{b\in\mathcal{V}}
\left(
\mathit{CS}_{\mathrm{structured}}^{(b,\mathrm{final})}
-
\mathit{CS}_{\mathrm{native}}^{(b,\mathrm{final})}
\right).
\end{equation}
Larger positive values indicate a greater increase in regional separation after rearrangement.

\paragraph{Cross-region Same-base Similarity.}
Cross-region Same-base Similarity ($C_{\mathrm{cross}}$) measures the similarity of same-nucleotide representations between the intervention interval and the background in the structured sequence:
\begin{equation}
C_{\mathrm{cross}}
=
\frac{1}{|\mathcal{V}|}
\sum_{b\in\mathcal{V}}
C_{\mathcal{I}\mathcal{B}}^{(b,\mathrm{final})}.
\end{equation}
Lower values indicate stronger regional separation of same-nucleotide representations.

\paragraph{Distal Representation Diffusion.}

Distal Representation Diffusion measures how a local rearrangement affects representations at unchanged positions outside the intervention interval. For each position $i\notin\mathcal{I}$, let $d_i$ denote its shortest distance to $\mathcal{I}$. The representation change and normalized distance are defined as
\begin{align}
D^{(\ell)}_{i}
&=
1-
\cos\left(
\mathbf{h}_{i,\mathrm{structured}}^{(\ell)},
\mathbf{h}_{i,\mathrm{native}}^{(\ell)}
\right),\\
r_i
&=
\frac{d_i}{d_{\max}},
\qquad
d_{\max}
=
\max_{j\notin\mathcal{I}}d_j.
\end{align}
Larger $D^{(\ell)}_{i}$ indicates greater context-induced representation change at position $i$, while $r_i$ provides a normalized positional distance for comparison across transcripts of different lengths.

Relative-distal Diffusion ($D_{\mathrm{distal}}$) averages the final-layer representation change over positions with $r_i\geq0.75$:
\begin{equation}
D_{\mathrm{distal}}
=
\mathbb{E}\left[
D^{(\mathrm{final})}_{i}
\mid
r_i\geq0.75
\right].
\end{equation}
The threshold $r_i \geq 0.75$ focuses on the most distal 25\% of positions. Lower values indicate weaker propagation into distant unchanged regions and are interpreted jointly with $\Delta\mathit{CS}$ and $C_{\mathrm{cross}}$.

\paragraph{Masked-nucleotide Accuracy.}
Masked-nucleotide Accuracy measures how well a model predicts nucleotides of the native transcript from their surrounding context. For a native transcript of length $\mathrm{L}$, a set $M$ of $\operatorname{round}(0.15\,\mathrm{L})$ positions is drawn once and shared across all models, and every position in $M$ is replaced by the mask token. The prediction at each masked position is the nucleotide with the highest logit among the four nucleotides in $\mathcal{V}$, and the accuracy is
\begin{equation}
\mathit{Acc}_{\mathrm{mask}}
=
\frac{1}{|M|}
\sum_{i\in M}
\mathbf{1}\!\left[
\underset{b\in\mathcal{V}}{\arg\max}\; z_i(b)=x_i
\right],
\end{equation}
where $z_i(b)$ is the logit assigned to nucleotide $b$ at position $i$. Restricting the prediction to the four nucleotides makes the measure comparable across models with different vocabularies. Higher values indicate better use of the sequence context. Because the length groups contain different transcripts, accuracy is compared across models within a length group rather than across groups.

All measures are first aggregated at the transcript-pair level and reported as means across 10 transcript pairs. Confidence intervals are estimated by bootstrap, and paired model comparisons report Cohen's $d_z$ and Benjamini-Hochberg-adjusted $q$-values (Appendix~\ref{app:statistical-analysis}).

\subsubsection{Long-Context Representation Analysis}

Beyond the pretrained 1K range, the evaluated models show three distinct behaviors (unrestricted values in Figure~\ref{fig:long-context-scaling}; Appendix Table~\ref{tab:cross-length-results}). First, most short-context models lose their representation organization under direct extrapolation, while YaRN partially restores this organization. Second, RiboGRAM, although also pretrained with a 1K context, remains stable up to 10,240~nt without any positional extension. Third, the models built for long inputs show no such failure: the native 10K \textsc{RiboSpan} variants, and HydraRNA, which serves as a hybrid state-space/attention reference to assess whether native dense attention provides an advantage in long-context representation modeling.

Under direct extrapolation, short-context dense Transformers progressively lose contextual organization beyond their pretrained range, with reduced $\Delta\mathit{CS}$ and increased $C_{\mathrm{cross}}$; the loss is most severe for RNAverse. Masked-nucleotide accuracy falls together with the representation metrics (Table~\ref{tab:long-context-results}). YaRN largely restores $\Delta\mathit{CS}$ and $C_{\mathrm{cross}}$ without changing the model weights, indicating that positional mismatch contributes substantially to this degradation. However, the recovery is accompanied by markedly increased $D_{\mathrm{distal}}$ and only a partial recovery of accuracy, suggesting that positional extension can restore context-dependent interactions without calibrating their propagation over transcript-scale distances.

\begin{table}[H]
\centering
\caption{Final-layer long-context representation metrics and masked-nucleotide accuracy at 10,240~nt. Bold and underlined values in result tables indicate the best and second-best performance, respectively.}
\label{tab:long-context-results}
\small
\begin{tabular}{llrrcr}
\toprule
Model & Setting & $\Delta\mathit{CS}\,\uparrow$ & $C_{\mathrm{cross}}\,\downarrow$ & $D_{\mathrm{distal}}$ & $\mathit{Acc}_{\mathrm{mask}}\,\uparrow$ \\
\midrule
HydraRNA~\citep{li2025hydrarna}
& Direct & 0.313846 & 0.521605 & 0.000667 & 0.427756 \\
\midrule
RiboGRAM~\citep{wang2026ribogram}
& Base & 0.417623 & 0.404581 & 0.002418 & 0.373698 \\
\midrule
RNAverse~\citep{ma2026rnaverse}
& Base & 0.066001 & 0.820777 & 0.004798 & 0.284632 \\
\midrule
\multirow{2}{*}{AIDO.RNA-CDS \citep{zou2024aido}}
& Base & 0.208178 & 0.660208 & 0.004666 & 0.334798 \\
& YaRN & \underline{0.446120} & 0.317370 & 0.025390 & 0.376911 \\
\midrule
\multirow{2}{*}{\textsc{RiboSpan}-1K-15}
& Base & 0.221320 & 0.698757 & 0.006626 & 0.309631 \\
& YaRN & \textbf{0.451068} & 0.334152 & 0.016084 & 0.392970 \\
\midrule
\multirow{2}{*}{\textsc{RiboSpan}-1K-40}
& Base & 0.186200 & 0.744354 & 0.004569 & 0.285818 \\
& YaRN & 0.406227 & 0.361375 & 0.029597 & 0.403019 \\
\midrule
\textsc{RiboSpan}-10K-15 & Native 10K & 0.405962 & \underline{0.302668} & 0.000785 & \textbf{0.608570} \\
\textsc{RiboSpan}-10K-40 & Native 10K & 0.405777 & \textbf{0.299277} & 0.001164 & \underline{0.605452} \\
\bottomrule
\end{tabular}
\end{table}

RiboGRAM and HydraRNA provide two architectural contrasts. RiboGRAM remains stable without positional extension, consistent with its parallel convolutional branches, which encode local sequence context independently of position and are therefore unaffected by positional extrapolation. This stability comes with weaker contextual differentiation: its $\Delta\mathit{CS}$ is comparable to that of the native 10K \textsc{RiboSpan} models, but its $C_{\mathrm{cross}}$ remains higher and its accuracy is among the lowest. HydraRNA is dominated by bidirectional state-space layers, with multi-head attention used in only two of twelve layers. Despite very small $D_{\mathrm{distal}}$, it shows substantially lower $\Delta\mathit{CS}$ and higher $C_{\mathrm{cross}}$ than the native 10K \textsc{RiboSpan} models, suggesting that its constrained propagation suppresses non-selective distal diffusion but also limits how strongly context differentiates the representations.

Together, these comparisons reveal complementary limitations: YaRN restores contextual differentiation but permits overly broad propagation, whereas RiboGRAM and HydraRNA remain stable but show weaker contextual differentiation. Native 10K pretraining is the only setting that combines strong contextual differentiation, localized perturbation responses, and the highest masked-nucleotide accuracy; the \textsc{RiboSpan}-1K variants, which share its corpus, architecture, and masking schedule, do not reach this accuracy under either direct extrapolation or YaRN.

These differences are most pronounced at 10,240~nt, while the qualitative behavior of $D_{\mathrm{distal}}$ remains robust to alternative distance thresholds (Appendix~\ref{app:diffusion-sensitivity}). The 10K-15 and 10K-40 checkpoints remain closely aligned in $\Delta\mathit{CS}$, $C_{\mathrm{cross}}$, and $\mathit{Acc}_{\mathrm{mask}}$, and both maintain very low $D_{\mathrm{distal}}$, indicating that the 40\% masking continuation largely preserves the representation profile established by native 10K pretraining.

\subsubsection{Attention Restriction Ablation}
\label{sec:attention-restriction}

To locate the source of the failure under direct extrapolation, we restrict every attention layer so that each position attends only within the model's pretrained context length, leaving model weights and position embeddings unchanged (Appendix~\ref{app:attention-restriction}). This restores the representation organization of the short-context models: $\Delta\mathit{CS}$ rises, $C_{\mathrm{cross}}$ falls, and $D_{\mathrm{distal}}$ drops to nearly zero, including under YaRN (Figure~\ref{fig:long-context-scaling}; Appendix Table~\ref{tab:attention-restriction}). For AIDO.RNA-CDS and the \textsc{RiboSpan}-1K variants, the restricted representation metrics reach or exceed those of the native 10K models. In contrast, the models that were already stable, namely RiboGRAM, HydraRNA, and the native 10K variants, change little. The failure therefore originates from attention to positions beyond the pretrained range rather than from the longer input itself.

However, restriction restores representation organization without recovering what native long-context pretraining provides. Under restriction, the masked-nucleotide accuracy of the short-context models improves but remains significantly below that of the native 10K variants at every length beyond the pretrained range (Appendix Table~\ref{tab:restricted-paired}). This holds for the \textsc{RiboSpan}-1K variants as well, which differ from the 10K variants only in native context length. Conversely, restricting the native 10K models to shorter spans leaves their representation metrics nearly unchanged but steadily lowers their accuracy, with a significant loss once the span falls to 2,048 tokens or fewer (Appendix Table~\ref{tab:span-sweep}). The same restriction to 1,024 tokens thus has opposite effects on the two model families, raising the accuracy of the 1K models and lowering that of the 10K models. Attention beyond the pretrained range disrupts short-context models, whereas native long-context pretraining makes it usable for prediction.

\begin{figure}[H]
    \centering
    \includegraphics[width=\textwidth]{figures/Long_Context_Benchmark.png}
    \caption{Length sweep of final-layer long-context representation metrics and masked-nucleotide accuracy. $\Delta\mathit{CS}$, $C_{\mathrm{cross}}$, $D_{\mathrm{distal}}$, and $\mathit{Acc}_{\mathrm{mask}}$ are shown across input lengths. Bars show the values under attention restriction, in which each position attends only to positions within the model's pretrained context length; hatched segments and open frames mark the increase or decrease relative to unrestricted attention. Error bars denote 95\% bootstrap confidence intervals.}
    \label{fig:long-context-scaling}
\end{figure}

\FloatBarrier

\newpage

\subsection{RNA Type Representation Benchmark}
\label{sec:rna-type-representation}

We evaluate RNA-type organization directly in the frozen representation space. By removing trainable downstream components, this benchmark provides a more direct and stringent assessment of the representations learned by the pretrained backbone itself. A strong pretrained encoder should therefore produce a representation space in which biologically related RNAs are already organized into locally coherent regions, allowing RNA identity to be recovered directly from the backbone representations without learned downstream adaptation.

\subsubsection{Benchmark Design}
\label{sec:representation-benchmark-design}

For each RNA sequence, final-layer hidden states over valid nucleotide positions are mean-pooled into a single sequence representation. For a sequence containing $L$ valid nucleotide tokens, the sequence representation is
\begin{equation}
\bar{h}
=
\frac{1}{L}\sum_{i=1}^{L} h_{\mathrm{final}}(i),
\end{equation}
where $h_{\mathrm{final}}(i)$ denotes the final-layer hidden state at nucleotide position $i$. All pretrained weights remain frozen, with no classifier, projection head, or downstream fine-tuning.

The Overall Biotype evaluation contains 89,955 sequences spanning 25 RNA biotypes with at least 20 examples per class. All sequences used in these evaluations are drawn exclusively from the held-out \textsc{RiboSpan} pretraining test set and were not used for \textsc{RiboSpan} pretraining optimization. Four focused label spaces further examine representation organization at different biological levels: Functional classes contain 60,892 sequences grouped into housekeeping, regulatory, and coding RNAs; Regulatory biotypes contain 29,895 sequences from seven regulatory RNA types; Long RNAs contain 17,130 sequences from eight biotypes with original sequence length greater than 1,024~nt; and the Rfam analysis contains 33,000 sequences from the 20 most frequent mapped Rfam families~\citep{ontiveros2025rfam}, providing an evaluation of conserved sequence and structural homology.

The evaluation includes the two \textsc{RiboSpan}-10K variants, with RNA-FM~\citep{chen2022rnafm}, RiNALMo~\citep{penic2025rinalmo}, AIDO.RNA-CDS~\citep{zou2024aido}, HydraRNA~\citep{li2025hydrarna}, RiboGRAM~\citep{wang2026ribogram}, and RNAverse~\citep{ma2026rnaverse} serving as external references. Inputs exceeding a model's effective context window are truncated to the supported length, while HydraRNA and \textsc{RiboSpan}-10K variants are evaluated on sequences up to 10,240~nt.

\subsubsection{Evaluation Measures}

Representation quality is evaluated using leave-one-out cosine $k$-nearest-neighbor label recovery with $k=10$. For each sequence representation $\bar{h}_i$, the 10 nearest neighbors are identified by cosine distance while excluding the query itself, and the predicted label $\hat{y}_i$ is assigned by majority vote. Voting ties are resolved deterministically by choosing the lexicographically smallest label. Label-recovery accuracy is
\begin{equation}
\mathit{Acc}
=
\frac{1}{n}\sum_{i=1}^{n}
\mathbf{1}\!\left[\hat{y}_i=y_i\right].
\end{equation}
Higher accuracy indicates that RNA identity can be more reliably recovered from the local geometric organization of the frozen representation space.

We additionally report Neighborhood Purity. For each sequence, the local purity $p_i$ is defined as the fraction of its $k$ nearest neighbors sharing the same label, and the reported purity is averaged over all sequences:
\begin{equation}
p_i
=
\frac{1}{k}\sum_{j\in\mathcal{N}_k(i)}
\mathbf{1}\!\left[y_j=y_i\right],
\qquad
\mathit{Purity}
=
\frac{1}{n}\sum_{i=1}^{n}p_i.
\end{equation}
Higher purity indicates stronger local concentration of same-type RNAs within the learned representation space and reduced mixing between different RNA classes.

\subsubsection{RNA Type Separability}

The frozen \textsc{RiboSpan} representations show strong RNA-type organization across the evaluated label spaces. In the Overall Biotype evaluation, \textsc{RiboSpan}-10K-15 achieves the highest accuracy and neighborhood purity among the models shown, with \textsc{RiboSpan}-10K-40 remaining closely matched (Table~\ref{tab:rna-type-representation}). This close agreement indicates that the reconstruction-oriented 40\% masking continuation preserves the overall RNA-type representation quality of the 10K pretrained checkpoint.

\begin{figure}[H]
    \centering
    \includegraphics[width=\textwidth]{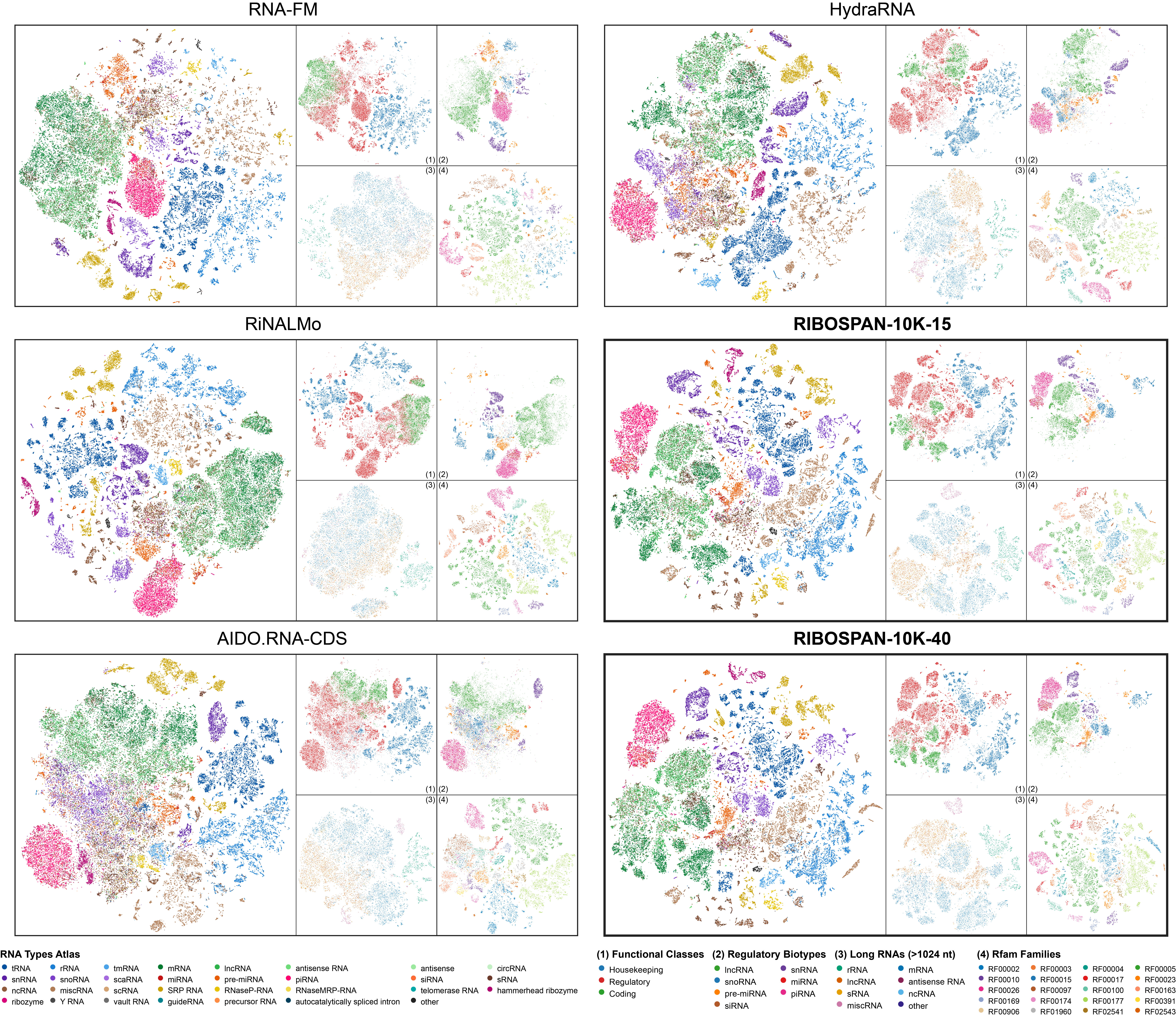}
    \caption{t-SNE visualization of frozen RNA sequence representations across Overall Biotype, Functional, Regulatory, Long RNA ($>1{,}024$~nt), and Rfam label spaces.}
    \label{fig:rna-type-tsne}
\end{figure}

At the coarse functional level, the two 10K checkpoints also achieve the highest accuracy and neighborhood purity among the models shown. Class-level analysis further shows strong organization across coding, housekeeping, and regulatory RNAs, together with reduced overall coding-regulatory confusion (Appendix~\ref{app:functional-results}). Both checkpoints also maintain strong separability across Regulatory biotypes.

A clear advantage of the \textsc{RiboSpan}-10K representations appears in the Long RNA evaluation. Both checkpoints achieve the highest accuracy and neighborhood purity among the models shown on sequences longer than 1,024~nt. Class-level analysis further shows that the \textsc{RiboSpan}-10K checkpoints achieve the strongest neighborhood purity in six of the eight long-RNA classes, including mRNA, lncRNA, miscRNA, and sRNA (Appendix~\ref{app:long-rna-results}), supporting strong representation organization across diverse long-RNA types.

\begin{table}[H]
\centering
\caption{RNA-type representation quality measured by leave-one-out 10-NN accuracy and neighborhood purity. Bold and underlined values in result tables indicate the best and second-best performance, respectively.}
\label{tab:rna-type-representation}
\small
\begin{tabular}{llccccc}
\toprule
Model & Metric & Overall Biotype & Functional & Regulatory & Long RNA & Rfam \\
\midrule
\multirow{2}{*}{RNA-FM~\citep{chen2022rnafm}}
& Acc.
& 0.865033
& 0.954608
& 0.953270
& 0.809924
& \textbf{0.990485} \\
& Purity
& 0.819314
& 0.939623
& 0.941489
& 0.727898
& \textbf{0.986791} \\
\midrule

\multirow{2}{*}{RiNALMo~\citep{penic2025rinalmo}}
& Acc.
& 0.857395
& 0.944081
& 0.952534
& 0.755867
& 0.985000 \\
& Purity
& 0.810520
& 0.927728
& 0.936635
& 0.673561
& 0.978312 \\
\midrule

\multirow{2}{*}{AIDO.RNA-CDS~\citep{zou2024aido}}
& Acc.
& 0.805125
& 0.954296
& 0.870781
& 0.841973
& 0.952242 \\
& Purity
& 0.735243
& 0.935233
& 0.813092
& 0.770362
& 0.936439 \\
\midrule

\multirow{2}{*}{HydraRNA~\citep{li2025hydrarna}}
& Acc.
& 0.861264
& 0.976237
& 0.925640
& 0.883771
& \underline{0.987606} \\
& Purity
& 0.811502
& 0.968406
& 0.894370
& 0.828219
& \underline{0.984218} \\
\midrule

\multirow{2}{*}{RiboGRAM~\citep{wang2026ribogram}}
& Acc.
& 0.862309
& 0.949074
& 0.953337
& 0.802393
& 0.980394 \\
& Purity
& 0.813478
& 0.932057
& 0.935340
& 0.719486
& 0.969879 \\
\midrule

\multirow{2}{*}{RNAverse~\citep{ma2026rnaverse}}
& Acc.
& 0.834539
& 0.938481
& 0.939890
& 0.750204
& 0.983939 \\
& Purity
& 0.777562
& 0.916608
& 0.919796
& 0.655254
& 0.978964 \\
\midrule

\multirow{2}{*}{\textsc{RiboSpan}-10K-15}
& Acc.
& \textbf{0.898861}
& \underline{0.980621}
& \textbf{0.959391}
& \underline{0.889375}
& 0.983818 \\
& Purity
& \textbf{0.859254}
& \underline{0.971592}
& \textbf{0.945871}
& \underline{0.836246}
& 0.974148 \\
\midrule

\multirow{2}{*}{\textsc{RiboSpan}-10K-40}
& Acc.
& \underline{0.898616}
& \textbf{0.980933}
& \underline{0.958087}
& \textbf{0.891010}
& 0.983970 \\
& Purity
& \underline{0.858702}
& \textbf{0.971944}
& \underline{0.944456}
& \textbf{0.836643}
& 0.975118 \\
\bottomrule
\end{tabular}
\end{table}

In the Rfam evaluation, RNA-FM achieves the strongest performance, while the \textsc{RiboSpan} checkpoints remain highly competitive, suggesting that \textsc{RiboSpan} captures distinct family-associated sequence and structural-homology patterns across Rfam families.

The t-SNE~\citep{vandermaaten2008tsne} visualizations further illustrate the organization of the frozen representation space (Figure~\ref{fig:rna-type-tsne}). Applied directly to the raw mean-pooled sequence representations without additional projection learning or downstream adaptation, t-SNE reveals clear and coherent RNA-type organization across all evaluated label spaces. Combined with the consistently strong accuracy and neighborhood purity across these evaluations, the results demonstrate that \textsc{RiboSpan} learns broadly transferable RNA representations and provides the strongest overall frozen representation quality among the evaluated models.

\medskip

\subsection{Downstream Biological Benchmarks}
\label{sec:downstream-benchmarks}

We consider three downstream settings: RNAGym~\citep{arora2025rnagym}, which evaluates zero-shot mutation fitness directly from pretrained sequence-model scores; mRNABench~\citep{shi2025mrnabench}, which evaluates biological information accessible from frozen mature-mRNA representations through linear probing; and RNA secondary structure prediction, which evaluates structural information under two published protocols.

\subsubsection{Zero-Shot Mutation Fitness Prediction on RNAGym}
\label{sec:rnagym}

We first evaluate zero-shot mutation fitness on RNAGym~\citep{arora2025rnagym} across diverse RNA categories. No task-specific predictor is trained. Instead, all bidirectional masked language models are evaluated using the same wild-type-background masked-marginal scoring procedure, and the resulting mutation scores are compared directly with experimentally measured fitness. Performance is assessed using Spearman correlation, AUROC, and MCC following the RNAGym evaluation procedure, with assay-level results first averaged within RNA category and then equally aggregated across categories. Detailed assay composition, scoring procedures, sequence windowing, and metric definitions are provided in Appendix~\ref{app:rnagym}.

As shown in Table~\ref{tab:rnagym-results}, all four \textsc{RiboSpan} checkpoints outperform the external encoder-only RNA language models across all three metrics under the unified masked-marginal evaluation. \textsc{RiboSpan}-1K-40 achieves the strongest overall performance and, in the broader RNAGym comparison, is surpassed only by Evo2-7B~\citep{brixi2026evo2}. These results establish \textsc{RiboSpan} as the strongest encoder-only RNA language model on RNAGym.

A consistent checkpoint-level pattern is also observed within \textsc{RiboSpan}. At both context lengths, the 40\% masking checkpoints outperform their 15\% counterparts, indicating that high-masking continuation further strengthens zero-shot mutation-fitness prediction. Both 10K checkpoints remain highly competitive and outperform all external encoder baselines, while the corresponding 1K checkpoints achieve even stronger performance. This pattern suggests that shorter native contexts may preserve finer-grained sensitivity to local nucleotide variation, providing an additional advantage for variant-centered prediction.

\begin{table}[H]
\centering
\caption{Zero-shot mutation-fitness performance on RNAGym. Bold and underlined values indicate the best and second-best performance \textbf{among encoder-only models}, respectively. Evo2-7B (reported by RNAGym~\citep{arora2025rnagym}) and EVA-1.4B are included as external decoder-only references.}
\label{tab:rnagym-results}
\small
\begin{tabular}{lcccc}
\toprule
Model & Model Type & Spearman $r$ & AUROC & MCC \\
\midrule
RNA-FM~\citep{chen2022rnafm}
& \multirow{7}{*}{Encoder-only (MLM)}
& 0.1674 & 0.5874 & 0.1279 \\
RiNALMo~\citep{penic2025rinalmo}
& & 0.2080 & 0.6051 & 0.1512 \\
AIDO.RNA-CDS~\citep{zou2024aido}
& & 0.1997 & 0.6024 & 0.1618 \\
HydraRNA~\citep{li2025hydrarna}
& & 0.1921 & 0.5989 & 0.1508 \\
RiboGRAM~\citep{wang2026ribogram}
& & 0.1958 & 0.6019 & 0.1544 \\
RNAverse~\citep{ma2026rnaverse}
& & 0.2132 & 0.6077 & 0.1608 \\
LIMBER-macro~\citep{wang2026limber}
& & 0.2105 & 0.6116 & 0.1640 \\
\midrule
\textsc{RiboSpan}-1K-15
& \multirow{4}{*}{Encoder-only (MLM)}
& \underline{0.2409} & \underline{0.6225} & \underline{0.1828} \\
\textsc{RiboSpan}-1K-40
& & \textbf{0.2524} & \textbf{0.6280} & \textbf{0.1904} \\
\textsc{RiboSpan}-10K-15
& & 0.2228 & 0.6169 & 0.1766 \\
\textsc{RiboSpan}-10K-40
& & 0.2301 & 0.6180 & 0.1820 \\
\midrule
Evo2-7B~\citep{brixi2026evo2}
& \multirow{2}{*}{Decoder-only (AR)}
& 0.2760 & 0.6360 & 0.2122 \\
EVA-1.4B~\citep{huang2026eva}
& & 0.1905 & 0.5971 & 0.1469 \\
\bottomrule
\end{tabular}
\end{table}

\subsubsection{Full-Transcript Property Prediction on mRNABench}
\label{sec:mrnabench}

We next evaluate frozen transcript-level representations across six mRNABench task families~\citep{shi2025mrnabench}, covering RNA half-life (HL), mean ribosome load (MRL), translation efficiency (TE), eCLIP binding, Gene Ontology annotations (GO), and variant-effect prediction (VEP). For each complete mature mRNA, final-layer nucleotide representations are mean-pooled over the full sequence. When the sequence exceeds a model's native context window, it is divided into non-overlapping native-window chunks whose pooled representations are combined by a length-weighted average, ensuring that every nucleotide contributes to the final transcript representation. All pretrained parameters remain frozen, and the resulting representations are evaluated using linear probes over fixed data splits. Detailed task composition, splitting procedures, probe settings, and metric definitions are provided in Appendix~\ref{app:mrnabench}.

\begin{table}[H]
\centering
\caption{Frozen full-transcript linear-probe performance on mRNABench. Bold and underlined values indicate the best and second-best performance, respectively.}
\label{tab:mrnabench-results}
\small
\begin{tabular}{lcccccc}
\toprule
Model & HL & MRL & TE & eCLIP & GO & VEP \\
Metric & Pearson $r$ & Pearson $r$ & Pearson $r$ & AUPRC & AUPRC & AUPRC \\
\midrule
RNA-FM~\citep{chen2022rnafm} & 0.4660 & 0.2928 & 0.6045 & 0.3507 & 0.3337 & 0.2835 \\
RiNALMo~\citep{penic2025rinalmo} & 0.4671 & 0.2802 & 0.6112 & 0.3613 & 0.3435 & 0.3253 \\
AIDO.RNA-CDS~\citep{zou2024aido} & 0.5613 & 0.3711 & 0.6769 & 0.4057 & 0.4449 & 0.3300 \\
HydraRNA~\citep{li2025hydrarna} & 0.6228 & 0.4605 & 0.7505 & 0.4611 & \textbf{0.4907} & 0.3376 \\
RiboGRAM~\citep{wang2026ribogram} & 0.4806 & 0.2713 & 0.6073 & 0.3523 & 0.3404 & 0.2903 \\
RNAverse~\citep{ma2026rnaverse} & 0.4891 & 0.3090 & 0.6416 & 0.3749 & 0.3585 & 0.3073 \\
LIMBER-macro~\citep{wang2026limber} & 0.5461 & 0.3666 & 0.6741 & 0.3929 & 0.3463 & 0.3468 \\
\midrule
\textsc{RiboSpan}-1K-15 & 0.6248 & 0.4637 & 0.7247 & 0.4374 & 0.4689 & \textbf{0.3480} \\
\textsc{RiboSpan}-1K-40 & 0.6128 & 0.4549 & 0.7150 & 0.4310 & 0.4660 & \underline{0.3473} \\
\textsc{RiboSpan}-10K-15 & \textbf{0.6522} & \textbf{0.4827} & \textbf{0.7610} & \textbf{0.4767} & \underline{0.4852} & 0.3430 \\
\textsc{RiboSpan}-10K-40 & \underline{0.6421} & \underline{0.4746} & \underline{0.7553} & \underline{0.4706} & 0.4824 & 0.3402 \\
\bottomrule
\end{tabular}
\end{table}

As shown in Table~\ref{tab:mrnabench-results}, \textsc{RiboSpan}-10K-15 shows the strongest overall performance, leading HL, MRL, TE, and eCLIP and ranking second in GO. The 10K-40 checkpoint follows a similar pattern, while VEP is the only task on which the 1K checkpoints outperform their 10K counterparts, with \textsc{RiboSpan}-1K-15 performing best.

When comparing checkpoints trained with the same masking rate, both 10K models outperform their 1K counterparts in five of the six task families, with VEP as the sole exception. This pattern echoes the preceding RNAGym benchmark, where the 1K checkpoints also perform better on variant-centered fitness prediction. Shorter native contexts may therefore retain finer-grained sensitivity to local nucleotide variation, whereas broader joint contextualization becomes increasingly important for phenotypes that depend on information distributed across the transcript. This distinction is particularly relevant to mRNABench, where most inputs exceed the 1K context and must be encoded as independent chunks. Native 10K modeling enables substantially broader cross-region interaction before pooling and consequently provides a consistent advantage across most transcript-level tasks.

Across RNAGym and mRNABench, context length and masking continuation reveal complementary effects. The 40\% continuation consistently improves zero-shot mutation-fitness prediction on RNAGym, whereas the 15\% checkpoints show stronger frozen-representation transfer on mRNABench. At the same time, the 1K checkpoints are favored for variant-centered prediction, while the 10K checkpoints show a broad advantage across transcript-level properties. Together, these results show that \textsc{RiboSpan} combines the strongest full-transcript transfer among the evaluated RNA encoders with the strongest zero-shot mutation-fitness performance among encoder-only RNA language models, while highlighting native long-context modeling as a particular advantage for biological phenotypes that require integration of distributed transcript-scale information.

\subsubsection{RNA Secondary Structure Prediction}
\label{sec:secondary-structure}

We further evaluate RNA secondary structure prediction, a task centered on non-coding RNAs, under two published protocols with the two \textsc{RiboSpan}-10K variants (Appendix~\ref{app:secondary-structure}). The structRFM protocol~\citep{zhu2026structrfm} trains a prediction head on bpRNA~\citep{danaee2018bprna} and on RNAStrAlign~\citep{tan2017turbofold} with evaluation on ArchiveII~\citep{sloma2016exact}; we keep the \textsc{RiboSpan} backbone frozen and train only the head, and compare against structRFM, a structure-guided RNA foundation model fine-tuned end to end. The LIMBER protocol~\citep{wang2026limber} fine-tunes each model and evaluates on bpRNA TS0, Rivas TestSetB~\citep{rivas2012range}, and ArchiveII under leave-one-family-out, a more demanding setting in which every sequence of the test family is withheld from training; the reference is LIMBER-macro, a model pretrained on non-coding RNAs.

Under the structRFM protocol, the frozen \textsc{RiboSpan} backbones outperform the fully fine-tuned structRFM on both ArchiveII and bpRNA TS0 (Figure~\ref{fig:secondary-structure}a; Appendix Table~\ref{tab:structrfm-protocol}). Under the LIMBER protocol, \textsc{RiboSpan} remains competitive on bpRNA TS0 and Rivas TestSetB, and on the leave-one-family-out evaluation it matches LIMBER-macro in pooled F1 and is higher in most of the nine held-out families (Figure~\ref{fig:secondary-structure}b; Appendix Table~\ref{tab:limber-protocol}). Six of these families contain pseudoknots in their reference structures; when crossing base pairs are scored alone, \textsc{RiboSpan}-10K-15 achieves the highest F1 in five of them and exceeds the dedicated pseudoknot predictor IPknot~\citep{sato2022ipknot} in all six (Appendix Table~\ref{tab:limber-pseudoknot}).

These results show that \textsc{RiboSpan}, although designed for full-length mRNA rather than specialized for non-coding RNA, is competitive with dedicated non-coding RNA models on structure prediction. They also indicate that explicit structural supervision during pretraining is not required: sequence-only pretraining at scale yields representations that already encode secondary structure.

\begin{figure}[H]
    \centering
    \includegraphics[width=\textwidth]{figures/Secondary_Structure_Benchmark.png}
    \caption{Per-sequence secondary structure prediction performance under (a) the structRFM protocol and (b) the LIMBER protocol. Violins show the test sets scored as a whole and boxes show ArchiveII by RNA family; lines and dots mark medians and means, and the number above each violin is the mean in (a) and the median in (b).}
    \label{fig:secondary-structure}
\end{figure}

\FloatBarrier


\section{Full-Length mRNA Generation Framework}
\label{sec:design}

Building on the pretrained \textsc{RiboSpan} backbone, we develop a conditional discrete-diffusion framework for full-length mRNA generation and sequence redesign, extending the bidirectional reconstruction capabilities learned during pretraining to generation. Unlike autoregressive models that decode nucleotides sequentially, the diffusion process jointly updates multiple positions along the denoising trajectory, allowing each step to integrate bidirectional context across the complete transcript. This formulation is well suited to full-length mRNA design, where the 5$^\prime$ UTR, CDS, and 3$^\prime$ UTR may jointly determine sequence properties. Dense bidirectional attention further allows each denoising step to integrate information across arbitrary transcript positions, while native 10K pretraining enables coordinated changes across distant regions.

\begin{figure}[H]
    \centering
    \includegraphics[width=0.85\textwidth]{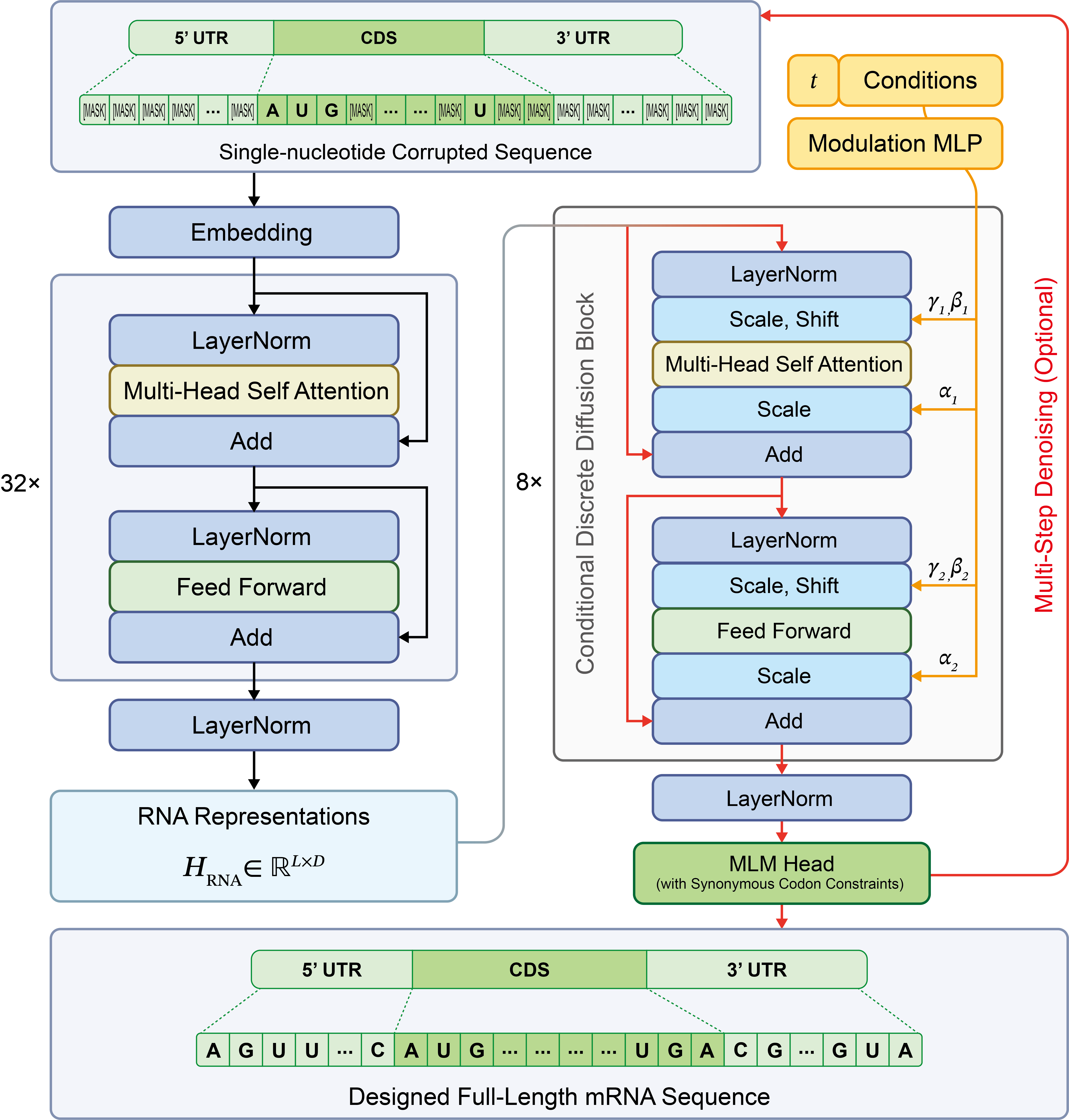}
    \caption{Overview of the \textsc{RiboSpan}-based conditional discrete-diffusion framework for full-length mRNA generation and redesign. Diffusion timesteps and design conditions are incorporated through AdaLN-Zero-conditioned blocks to guide nucleotide denoising, with synonymous-codon constraints for CDS-preserving optimization.}
    \label{fig:generation-framework}
\end{figure}

At each diffusion step, the corrupted single-nucleotide sequence is first encoded by the pretrained \textsc{RiboSpan} backbone to obtain contextualized RNA representations. The diffusion timestep $t$ and multidimensional design conditions $c$ are then combined through a modulation MLP and injected into each conditional diffusion block through AdaLN-Zero conditioning~\citep{peebles2023scalable} (Figure~\ref{fig:generation-framework}). The modulation network produces layer-specific scale, shift, and residual-gating parameters, allowing the design conditions to continuously modulate the hidden representations throughout the denoising trajectory. After the conditional diffusion blocks, an MLM head predicts nucleotide distributions for reconstruction of the clean sequence.

The generation module is trained using a conditional masked-diffusion objective inspired by MDLM~\citep{sahoo2024simple}. For a clean sequence $x$ containing $L$ valid nucleotide positions and a design condition $c$, a diffusion timestep $t$ determines the corruption level, and a corresponding set of positions $M_t$ is replaced by mask tokens to form the corrupted sequence $z_t$. Under the linear masking schedule used here, the continuous-time masked-diffusion objective gives
\begin{equation}
\mathcal{L}_{\mathrm{diff}}
=
\mathbb{E}_{t}
\left[
\frac{1}{tL}
\sum_{i\in M_t}
-\log p_{\theta}\!\left(x_i\mid z_t,c,t\right)
\right].
\end{equation}
During training, stratified timestep sampling distributes corrupted examples across the diffusion horizon, allowing the model to learn sequence recovery from different levels of corruption.

The framework supports flexible full-length mRNA design while preserving joint modeling of sequence context across the 5$'$ UTR, CDS, and 3$'$ UTR. Because all designable positions are updated within the same bidirectional long-context representation, nucleotide changes in one region are generated in the context of the entire transcript, allowing the model to account for long-range dependencies and coordinated sequence patterns across distant mRNA regions. CDS design is implemented through synonymous-codon diffusion, which restricts codon substitutions to synonymous alternatives to preserve the encoded amino-acid sequence while allowing coding and noncoding regions to be jointly optimized under full-transcript context. These capabilities enable \textit{de novo} generation, full-length sequence redesign, and cross-region constrained optimization within a unified generative framework.

Beyond the conditional diffusion framework described here, the broader \textsc{RiboSpan} design system integrates RNA property prediction with reinforcement-learning post-training to enable closed-loop sequence optimization. Task-specific predictors provide quantitative feedback on generated sequences, which can be incorporated into the reinforcement-learning objective together with biological constraints to steer generation toward desired multidimensional functional profiles. The complete mRNA design and reinforcement-learning post-training framework, including the corresponding model checkpoints, will be presented in a forthcoming journal publication together with experimental evaluation of the biological performance and downstream applications of the designed sequences.


\section{Conclusion}
\label{sec:conclusion}

We introduced \textsc{RiboSpan}, a 1.61B-parameter bidirectional RNA foundation model with single-nucleotide tokenization, dense self-attention, and native pretraining at context lengths up to 10,240~nt. Trained on 67.6 million RNA sequences comprising 85.7 billion nucleotide tokens, \textsc{RiboSpan} extends high-resolution bidirectional representation learning to long RNAs and full-length mRNAs while learning transferable representations across diverse RNA types. Across reconstruction and long-context evaluations, native 10K pretraining provides clear advantages over extending short-context models beyond their pretrained range. The native 10K variants retain strong reconstruction at 10,240 tokens, while the reconstruction-oriented 40\% masking continuation improves recovery under heavy corruption and preserves representation quality. Our long-context benchmark shows that native 10K pretraining combines strong contextual differentiation, localized perturbation responses, and the highest masked-nucleotide accuracy, whereas direct extrapolation, inference-time positional scaling, and alternative architectures each fall short in at least one of these respects. An attention-restriction ablation further traces the failure of short-context models to attention beyond the pretrained range, and shows that restricting attention restores representation organization but not the accuracy of native long-context pretraining. Frozen-representation evaluations further demonstrate state-of-the-art RNA representation quality, with the strongest overall performance and a clear advantage on long RNAs while maintaining strong organization across diverse RNA types and functional label spaces. Across downstream biological benchmarks, \textsc{RiboSpan} further emerges as the strongest encoder-only RNA foundation model, achieving state-of-the-art performance in full-transcript biological property prediction and zero-shot mutation-fitness modeling. On RNA secondary structure prediction, \textsc{RiboSpan} outperforms a fully fine-tuned structure-guided model with its backbone frozen and is competitive with a dedicated non-coding RNA model, indicating that sequence-only pretraining at scale suffices to encode structural information. Building on the same backbone, we developed a conditional discrete-diffusion framework for full-length mRNA generation and redesign, jointly modeling the 5$'$ UTR, CDS, and 3$'$ UTR under complete-transcript context and enabling protein-preserving CDS optimization through synonymous-codon diffusion. \textsc{RiboSpan} further integrates RNA property prediction and reinforcement-learning post-training toward closed-loop multidimensional sequence optimization. Together, \textsc{RiboSpan} unifies high-resolution long-context RNA representation learning, broad biological prediction, and full-transcript generative modeling, establishing a versatile foundation for RNA understanding and cross-region, multi-objective mRNA design.

\clearpage


\clearpage
\appendix


\section{Pretraining Details}
\label{app:pretraining}
\label{app:training-details}

\subsection{Data Curation and Splits}
\label{app:data-curation}

RNAcentral v26.0, Ensembl release 115, and Ensembl Genomes release 62 were curated independently before merging. All sequences were normalized to uppercase, with U mapped to T and unsupported symbols mapped to N. For RNAcentral, the active and inactive sequence sets were combined and exact-deduplicated by sequence using SeqKit~\citep{shen2016seqkit}. Ensembl cDNA sequences from vertebrates, plants, fungi, metazoans, and protists were paired with the corresponding release-matched non-ab-initio GTF annotations and filtered to retain complete, high-quality protein-coding transcripts according to Table~\ref{tab:ensembl-filtering}, followed by sequence-level exact deduplication using SeqKit. Table~\ref{tab:corpus-source-stats} summarizes the resulting source-level sequence and nucleotide-token counts.

\begin{table}[H]
\centering
\caption{Pretraining corpus statistics by split and source.}
\label{tab:corpus-source-stats}
\small
\begin{tabular}{llrr}
\toprule
Split & Source & Sequences & Nucleotide Tokens \\
\midrule
Training & RNAcentral v26.0 & 49,250,975 & 34,111,085,735 \\
Training & Ensembl 115 \& Ensembl Genomes 62& 18,316,110 & 51,603,244,706 \\
Validation & Combined & 90,000 & 77,781,035 \\
Test & Combined & 90,000 & 76,893,030 \\
\bottomrule
\end{tabular}
\end{table}

\begin{table}[H]
\centering
\caption{Filtering criteria for Ensembl protein-coding transcripts.}
\label{tab:ensembl-filtering}
\small
\begin{tabular}{lp{0.72\linewidth}}
\toprule
Criterion & Requirement \\
\midrule
Annotation & Protein-coding transcript annotation and, when available, protein-coding gene annotation; neither is annotated as a pseudogene. \\
Completeness & CDS, start codon, stop codon, 5$'$ UTR, and 3$'$ UTR annotations are present. \\
Coding consistency & CDS length is positive and divisible by 3, with consistent start- and stop-codon annotations. \\
Sequence quality & $\geq 20$~nt, $\leq 5\%$ N, and no homopolymer $>50$~nt. \\
\bottomrule
\end{tabular}
\end{table}

\subsection{Tokenization and Training Sample Construction}
\label{app:training-samples}

Pretraining uses single-nucleotide tokenization over the normalized A/C/G/T/N sequences. Each RNA is retained as an independent sequence document. During sample construction, a \texttt{[CLS]} token and a \texttt{[SEP]} token are added to the beginning and end of each sequence, respectively, and both count toward the model context length. RNAs exceeding the native context length are truncated to fit the corresponding context window.

For masked language modeling, 80\% of selected positions are replaced by \texttt{[MASK]}, 10\% by a random token, and 10\% remain unchanged. Selected positions may form spans of up to three nucleotides. The same corruption procedure is used for the 15\% pretraining stage and the 40\% continuation.

To reduce padding overhead, sequences within each microbatch are ordered by valid length and greedily packed into bins bounded by the native context length. Transformer Engine packed attention preserves independent attention boundaries for each RNA, with positional indices reset to zero at sequence boundaries.

\subsection{Optimization Schedule}
\label{app:optimization}

For both context-length branches, the 40\% continuation resumes from the corresponding 15\% checkpoint with a newly initialized optimizer. Table~\ref{tab:optimization-settings} summarizes the training hyperparameters for both stages.

\begin{table}[H]
\centering
\caption{Hyperparameters for \textsc{RiboSpan} pretraining.}
\label{tab:optimization-settings}
\small
\begin{tabular}{lcc}
\toprule
Setting & 15\% MLM & 40\% MLM Continuation \\
\midrule
Epochs & 6 & 2 \\
Global Batch Size & 2,048 sequences & 2,048 sequences \\
Optimizer & AdamW & AdamW \\
Peak LR & $5\times10^{-5}$ & $1\times10^{-5}$ \\
Min LR & $1\times10^{-5}$ & $1\times10^{-6}$ \\
LR Scheduler & cosine & cosine \\
Warmup & 2,000 steps & 2,000 steps \\
Weight Decay & 0.01 & 0.01 \\
Clip Norm & 1.0 & 1.0 \\
Dropout (hidden / attn.) & 0.0 / 0.1 & 0.0 / 0.1 \\
Precision & BF16 & BF16 \\
\bottomrule
\end{tabular}
\end{table}

\FloatBarrier


\section{Long-Context Representation Benchmark Details}
\label{app:long-context}

\subsection{Long-Sequence Evaluation Settings}
\label{app:position-extension}

Model abbreviations follow the main text. HydraRNA is evaluated directly at each requested sequence length. AIDO.RNA-CDS, RiboGRAM, RNAverse, and the \textsc{RiboSpan}-1K variants are evaluated beyond 1,024 tokens by direct RoPE extrapolation. The suffix \texttt{-YaRN} denotes inference-time YaRN scaling, which extends the usable context range of short-context RoPE through frequency-dependent rotary interpolation and attention-score rescaling, without additional training.

Let $L$ denote the input sequence length and $L_0=1024$ the pretrained context length of the short-context models. Following YaRN~\citep{peng2024yarn}, the dynamic context-extension factor is
\begin{equation}
s(L)=\max\left(1,\frac{L}{L_0}\right).
\end{equation}
For RoPE dimension $d$ with angular frequency $\theta_d$ and wavelength $\lambda_d=2\pi/\theta_d$, define the number of rotations within the pretrained context as
\begin{equation}
r_d=\frac{L_0}{\lambda_d}.
\end{equation}
YaRN applies NTK-by-parts interpolation using
\begin{equation}
\gamma(r)=
\begin{cases}
0, & r<\alpha,\\
1, & r>\beta,\\
\dfrac{r-\alpha}{\beta-\alpha}, & \text{otherwise},
\end{cases}
\end{equation}
and modifies each rotary frequency as
\begin{equation}
\widetilde{\theta}_d
=
\left(1-\gamma(r_d)\right)\frac{\theta_d}{s}
+
\gamma(r_d)\theta_d.
\end{equation}
We use rotation-count thresholds $\alpha=1$ and $\beta=4$ to determine the transition between interpolated and unmodified rotary frequencies. YaRN additionally rescales the attention logits using the temperature
\begin{equation}
t(s)=\frac{1}{\left(0.1\ln s+1\right)^2}.
\end{equation}
Direct RoPE extrapolation uses the original rotary frequencies without positional rescaling, whereas the YaRN configurations apply the frequency interpolation and attention scaling above with unchanged model weights.

\subsection{Benchmark Construction and Paired Intervention}
\label{app:benchmark-panel}

Candidate mRNAs are restricted to unambiguous A/C/G/T/U sequences and normalized from U to T before sampling. For each target length $L_t\in\{1024,2048,4096,8192,10240\}$, transcripts satisfying
\begin{equation}
|L-L_t|\leq 0.01L_t
\end{equation}
are eligible. 10 complete transcripts are deterministically sampled without replacement from each length group, yielding 50 transcripts in total. All model configurations are evaluated on the same transcript panel, without cropping, padding, or concatenation.

For each transcript $x=(x_1,\ldots,x_L)$, the paired intervention uses the centered interval $\mathcal{I}$ of width $W=\operatorname{round}(L/32)$ defined in Section~\ref{sec:benchmark-design}. Let $i_1<\cdots<i_W$ denote the positions in $\mathcal{I}$. For each nucleotide $b\in\mathcal{V}$, where $\mathcal{V}=\{\mathrm{A},\mathrm{C},\mathrm{G},\mathrm{T}\}$ as in the main text, its count within the native interval is
\begin{equation}
n_b
=
\sum_{j=1}^{W}
\mathbf{1}\!\left[x_{i_j}=b\right].
\end{equation}
The candidate set $\mathcal{C}$ is constructed by permuting the nonempty nucleotide blocks $b^{n_b}$, yielding at most $4!=24$ candidates, each with exactly the same nucleotide composition as the native interval.

For a candidate $z=(z_1,\ldots,z_W)\in\mathcal{C}$, define the adjacent transition counts
\begin{equation}
c_{ab}(z)
=
\sum_{j=1}^{W-1}
\mathbf{1}\!\left[z_j=a,\;z_{j+1}=b\right],
\qquad
c_a(z)
=
\sum_{b\in\mathcal{V}}c_{ab}(z).
\end{equation}
The first-order transition conditional entropy, measured in bits, is
\begin{equation}
H_{\mathrm{tr}}(z)
=
-\sum_{\substack{a\in\mathcal{V}\\c_a(z)>0}}
\frac{c_a(z)}{W-1}
\sum_{\substack{b\in\mathcal{V}\\c_{ab}(z)>0}}
\frac{c_{ab}(z)}{c_a(z)}
\log_2\frac{c_{ab}(z)}{c_a(z)}.
\end{equation}
The Hamming distance from the native interval is
\begin{equation}
d_{\mathrm{H}}(z)
=
\sum_{j=1}^{W}
\mathbf{1}\!\left[z_j\neq x_{i_j}\right].
\end{equation}
To quantify short-period repetition, the identity at lag $\ell$ is
\begin{equation}
r_\ell(z)
=
\frac{1}{W-\ell}
\sum_{j=1}^{W-\ell}
\mathbf{1}\!\left[z_j=z_{j+\ell}\right],
\end{equation}
and the corresponding short-period score is
\begin{equation}
R_{\mathrm{short}}(z)
=
\max_{1\leq \ell\leq \min(12,\lfloor W/2\rfloor)}
r_\ell(z).
\end{equation}

Excluding the native arrangement whenever a distinct candidate exists, the structured interval is selected by lexicographically minimizing
\begin{equation}
z^\star
=
\arg\min_{z\in\mathcal{C}}
\left(
H_{\mathrm{tr}}(z),
-d_{\mathrm{H}}(z),
-R_{\mathrm{short}}(z)
\right).
\end{equation}
Exact ties are resolved by a deterministic, pair-specific permutation of candidate order. The structured sequence is obtained by replacing $x_{i_j}$ with $z^\star_j$ for $j=1,\ldots,W$, leaving all positions outside $\mathcal{I}$ unchanged. This construction preserves transcript length, interval nucleotide composition, and the surrounding sequence context.

\FloatBarrier

\subsection{Endpoint Sampling and Aggregation}
\label{app:endpoint-sampling}

The three primary representation endpoints defined in the main text are computed from final-layer hidden states, with boundary-token positions excluded. For Additional Context Separation and Cross-region Same-base Similarity, at most 512 positions are sampled for each nucleotide type and region, and each within-interval, within-background, and cross-region similarity estimate uses at most 512 position pairs. Sampling coordinates are determined independently of the model configuration and reused across all models. The pair-count weighting of within-region similarities follows the definition in the main text, and nucleotide types with valid estimates are averaged equally.

For Distal Representation Diffusion, positions whose input nucleotide differs between the native and structured sequences are excluded, so representation change is evaluated only at unchanged coordinates. Unchanged positions are sampled over the absolute-distance ranges $[0,8)$, $[8,32)$, $[32,128)$, $[128,512)$, $[512,1024)$, $[1024,2048)$, $[2048,4096)$, and $[4096,10240)$, with at most 1,024 positions retained per range. The same sampled coordinates are used across model configurations. Relative-distal Diffusion is computed from the retained positions satisfying $r_i\geq0.75$, using the normalized-distance and representation-change definitions in the main text.

\subsection{Statistical Analysis}
\label{app:statistical-analysis}

For the transcript-pair-level endpoint values described in the main text, 95\% confidence intervals are estimated using 2,000 bootstrap resamples within each length group.

For a given endpoint, let $y_{A,k}$ and $y_{B,k}$ denote the values obtained by models $A$ and $B$, respectively, on the $k$th matched transcript pair. The paired difference is
\begin{equation}
\delta_k
=
y_{B,k}-y_{A,k},
\qquad
k=1,\ldots,10.
\end{equation}
Paired Cohen's $d_z$ is defined as
\begin{equation}
d_z
=
\frac{\bar{\delta}}{s_{\delta}},
\end{equation}
where $\bar{\delta}$ and $s_{\delta}$ are the mean and standard deviation of the ten paired differences.

Two-sided bootstrap sign $p$-values are computed from the bootstrap distribution of $\bar{\delta}$ with finite-resampling correction. Multiple comparisons are controlled using the Benjamini-Hochberg procedure across the comparison family. Adjusted $q$-values below 0.002 are reported as $q<0.002$.

\subsection{Cross-Length Representation Results}
\label{app:cross-length-results}

Table~\ref{tab:cross-length-results} summarizes the three representation metrics and masked-nucleotide accuracy across input lengths from 1,024 to 8,192~nt, providing a cross-length view of how contextual differentiation, distal propagation, and accuracy evolve with increasing sequence length. The corresponding 10,240-nt results are reported separately in Table~\ref{tab:long-context-results}.

\begin{table}[H]
\centering
\caption{Long-context representation metrics and masked-nucleotide accuracy at 1,024--8,192~nt.}
\label{tab:cross-length-results}
\footnotesize
\setlength{\tabcolsep}{3pt}
\begin{adjustbox}{max width=\linewidth}
\begin{tabular}{@{}lcccc@{\hspace{1.2em}}lcccc@{}}
\toprule
\multicolumn{5}{@{}l}{\textit{Panel A. Additional Context Separation ($\Delta\mathit{CS}\,\uparrow$)}} &
\multicolumn{5}{@{\hspace{1.2em}}l@{}}{\textit{Panel C. Relative-distal Diffusion ($D_{\mathrm{distal}}$)}} \\
\addlinespace[2pt]
Model & 1,024 & 2,048 & 4,096 & 8,192 &
Model & 1,024 & 2,048 & 4,096 & 8,192 \\
\cmidrule(r){1-5}\cmidrule(l){6-10}
HydraRNA      & 0.034261 & 0.132366 & 0.258054 & 0.260502 & HydraRNA      & 0.002292 & 0.001554 & 0.000762 & 0.000518 \\
RiboGRAM      & 0.117905 & 0.246031 & 0.366716 & 0.397198 & RiboGRAM      & 0.002372 & 0.001980 & 0.003393 & 0.002762 \\
RNAverse      & 0.205579 & 0.275696 & 0.149930 & 0.082901 & RNAverse      & 0.001955 & 0.004005 & 0.007299 & 0.004232 \\
AIDO-CDS      & 0.042053 & 0.209758 & 0.385709 & 0.295581 & AIDO-CDS      & 0.006310 & 0.004948 & 0.006495 & 0.006473 \\
AIDO-CDS-YaRN & 0.041869 & 0.200108 & 0.375539 & 0.406480 & AIDO-CDS-YaRN & 0.006373 & 0.005309 & 0.017378 & 0.016653 \\
1K-15         & 0.061452 & 0.145217 & 0.327177 & 0.234125 & 1K-15         & 0.005218 & 0.003451 & 0.005900 & 0.006270 \\
1K-15-YaRN    & 0.061353 & 0.155104 & 0.325023 & 0.374779 & 1K-15-YaRN    & 0.005247 & 0.003720 & 0.005285 & 0.011854 \\
1K-40         & 0.039375 & 0.164191 & 0.244088 & 0.217073 & 1K-40         & 0.006681 & 0.007931 & 0.009690 & 0.005400 \\
1K-40-YaRN    & 0.039466 & 0.148706 & 0.301319 & 0.365205 & 1K-40-YaRN    & 0.006713 & 0.011811 & 0.010901 & 0.026445 \\
10K-15        & 0.060465 & 0.143069 & 0.332512 & 0.349363 & 10K-15        & 0.006556 & 0.002062 & 0.001261 & 0.000719 \\
10K-40        & 0.045734 & 0.142012 & 0.331983 & 0.343966 & 10K-40        & 0.006828 & 0.002311 & 0.001580 & 0.000960 \\
\midrule
\multicolumn{5}{@{}l}{\textit{Panel B. Cross-region Same-base Similarity ($C_{\mathrm{cross}}\,\downarrow$)}} &
\multicolumn{5}{@{\hspace{1.2em}}l@{}}{\textit{Panel D. Masked-nucleotide Accuracy ($\mathit{Acc}_{\mathrm{mask}}\,\uparrow$)}} \\
\addlinespace[2pt]
Model & 1,024 & 2,048 & 4,096 & 8,192 &
Model & 1,024 & 2,048 & 4,096 & 8,192 \\
\cmidrule(r){1-5}\cmidrule(l){6-10}
HydraRNA      & 0.709078 & 0.603054 & 0.528817 & 0.532895 & HydraRNA      & 0.443488 & 0.467861 & 0.479379 & 0.467838 \\
RiboGRAM      & 0.508257 & 0.451191 & 0.412732 & 0.412882 & RiboGRAM      & 0.400591 & 0.402039 & 0.387651 & 0.349769 \\
RNAverse      & 0.550082 & 0.576254 & 0.752767 & 0.822265 & RNAverse      & 0.361255 & 0.329102 & 0.299115 & 0.278892 \\
AIDO-CDS      & 0.485169 & 0.372824 & 0.373184 & 0.565919 & AIDO-CDS      & 0.445483 & 0.470129 & 0.432591 & 0.331470 \\
AIDO-CDS-YaRN & 0.485266 & 0.371601 & 0.313101 & 0.349170 & AIDO-CDS-YaRN & 0.444829 & 0.465861 & 0.425833 & 0.382420 \\
1K-15         & 0.467053 & 0.349655 & 0.498187 & 0.679041 & 1K-15         & 0.583843 & 0.628206 & 0.494537 & 0.305064 \\
1K-15-YaRN    & 0.467240 & 0.326959 & 0.320326 & 0.369014 & 1K-15-YaRN    & 0.581895 & 0.615800 & 0.613608 & 0.417887 \\
1K-40         & 0.499650 & 0.382734 & 0.635389 & 0.702894 & 1K-40         & 0.577931 & 0.623638 & 0.406951 & 0.285719 \\
1K-40-YaRN    & 0.499737 & 0.346924 & 0.335873 & 0.382344 & 1K-40-YaRN    & 0.577286 & 0.603665 & 0.607382 & 0.415132 \\
10K-15        & 0.492230 & 0.357921 & 0.314809 & 0.328785 & 10K-15        & 0.573904 & 0.698564 & 0.765558 & 0.587740 \\
10K-40        & 0.520815 & 0.362487 & 0.312377 & 0.329731 & 10K-40        & 0.576565 & 0.695626 & 0.762790 & 0.584079 \\
\bottomrule
\end{tabular}
\end{adjustbox}
\end{table}

\subsection{Relative-distal Diffusion Threshold Sensitivity}
\label{app:diffusion-sensitivity}

The primary Relative-distal Diffusion endpoint uses $r_i \geq 0.75$ to characterize representation change in the most distal 25\% of the normalized distance range. We additionally evaluate thresholds of $r_i \geq 0.25$ and $r_i \geq 0.50$. The same qualitative patterns persist across thresholds: YaRN produces substantially broader distal changes, whereas HydraRNA and the native 10K models remain strongly localized.

\begin{table}[H]
\centering
\caption{Sensitivity of Relative-distal Diffusion to the normalized-distance threshold at 10,240~nt.}
\label{tab:distal-sensitivity}
\small
\begin{tabular}{lccc}
\toprule
Model & $r_i \geq 0.25$ & $r_i \geq 0.50$ & $r_i \geq 0.75$ \\
\midrule
HydraRNA & 0.000636 & 0.000632 & 0.000667 \\
RiboGRAM & 0.002166 & 0.002313 & 0.002418 \\
RNAverse & 0.004851 & 0.004694 & 0.004798 \\
AIDO-CDS & 0.004527 & 0.004939 & 0.004666 \\
AIDO-CDS-YaRN & 0.030155 & 0.027632 & 0.025390 \\
1K-15 & 0.006011 & 0.006095 & 0.006626 \\
1K-15-YaRN & 0.018542 & 0.016600 & 0.016084 \\
1K-40 & 0.004245 & 0.004344 & 0.004569 \\
1K-40-YaRN & 0.029908 & 0.030132 & 0.029597 \\
10K-15 & 0.000867 & 0.000770 & 0.000785 \\
10K-40 & 0.001356 & 0.001234 & 0.001164 \\
\bottomrule
\end{tabular}
\end{table}

\subsection{Paired Model Comparisons at 10,240 nt}
\label{app:paired-comparisons}

Table~\ref{tab:paired-comparisons-10240} reports paired comparisons at 10,240~nt for the three primary representation endpoints and masked-nucleotide accuracy. For each comparison, $\Delta$ is defined as the value for Model B minus that for Model A.

\begin{table}[H]
\centering
\caption{Paired model comparisons at 10,240~nt.}
\label{tab:paired-comparisons-10240}
\small
\begin{tabular}{llrccr}
\toprule
Model A & Model B & $\Delta$ (B$-$A) & 95\% CI & $d_z$ & BH $q$ \\
\midrule
\multicolumn{6}{l}{\textit{Panel A. Additional Context Separation ($\Delta\mathit{CS}\uparrow$)}} \\
AIDO-CDS & AIDO-CDS-YaRN & 0.237942 & $[0.221561,\,0.252913]$ & 9.011 & $<0.002$ \\
AIDO-CDS-YaRN & 10K-15 & -0.040159 & $[-0.056615,\,-0.024325]$ & -1.472 & $<0.002$ \\
AIDO-CDS-YaRN & HydraRNA & -0.132275 & $[-0.148514,\,-0.115630]$ & -4.645 & $<0.002$ \\
HydraRNA & 10K-15 & 0.092116 & $[0.068073,\,0.115978]$ & 2.199 & $<0.002$ \\
RiboGRAM & 10K-15 & -0.011661 & $[-0.027754,\,0.005000]$ & -0.408 & 0.202466 \\
RNAverse & 10K-15 & 0.339961 & $[0.320451,\,0.361014]$ & 9.482 & $<0.002$ \\
1K-15 & 1K-15-YaRN & 0.229748 & $[0.202679,\,0.257008]$ & 4.966 & $<0.002$ \\
1K-15 & 10K-15 & 0.184642 & $[0.161994,\,0.206427]$ & 4.921 & $<0.002$ \\
1K-40 & 1K-40-YaRN & 0.220027 & $[0.185046,\,0.254946]$ & 3.687 & $<0.002$ \\
1K-40 & 10K-40 & 0.219577 & $[0.196458,\,0.242354]$ & 5.545 & $<0.002$ \\
10K-15 & 10K-40 & -0.000185 & $[-0.005737,\,0.005178]$ & -0.019 & 0.975611 \\
\midrule
\multicolumn{6}{l}{\textit{Panel B. Cross-region Same-base Similarity ($C_{\mathrm{cross}}\downarrow$)}} \\
AIDO-CDS & AIDO-CDS-YaRN & -0.342838 & $[-0.358145,\,-0.327690]$ & -12.989 & $<0.002$ \\
AIDO-CDS-YaRN & 10K-15 & -0.014702 & $[-0.023312,\,-0.005753]$ & -1.012 & 0.004042 \\
AIDO-CDS-YaRN & HydraRNA & 0.204234 & $[0.190618,\,0.218636]$ & 8.758 & $<0.002$ \\
HydraRNA & 10K-15 & -0.218937 & $[-0.234161,\,-0.205861]$ & -9.143 & $<0.002$ \\
RiboGRAM & 10K-15 & -0.101913 & $[-0.116755,\,-0.087660]$ & -4.214 & $<0.002$ \\
RNAverse & 10K-15 & -0.518109 & $[-0.532743,\,-0.500252]$ & -18.842 & $<0.002$ \\
1K-15 & 1K-15-YaRN & -0.364605 & $[-0.387405,\,-0.338777]$ & -8.812 & $<0.002$ \\
1K-15 & 10K-15 & -0.396089 & $[-0.414138,\,-0.374275]$ & -11.695 & $<0.002$ \\
1K-40 & 1K-40-YaRN & -0.382979 & $[-0.414869,\,-0.349068]$ & -6.856 & $<0.002$ \\
1K-40 & 10K-40 & -0.445077 & $[-0.469631,\,-0.414508]$ & -9.617 & $<0.002$ \\
10K-15 & 10K-40 & -0.003391 & $[-0.007703,\,0.000891]$ & -0.462 & 0.140955 \\
\midrule
\multicolumn{6}{l}{\textit{Panel C. Relative-distal Diffusion ($D_{\mathrm{distal}}$)}} \\
AIDO-CDS & AIDO-CDS-YaRN & 0.020724 & $[0.014688,\,0.026750]$ & 2.021 & $<0.002$ \\
AIDO-CDS-YaRN & 10K-15 & -0.024604 & $[-0.030803,\,-0.018826]$ & -2.420 & $<0.002$ \\
AIDO-CDS-YaRN & HydraRNA & -0.024723 & $[-0.030972,\,-0.019237]$ & -2.411 & $<0.002$ \\
HydraRNA & 10K-15 & 0.000118 & $[-0.000219,\,0.000415]$ & 0.224 & 0.505468 \\
RiboGRAM & 10K-15 & -0.001633 & $[-0.002042,\,-0.001224]$ & -2.414 & $<0.002$ \\
RNAverse & 10K-15 & -0.004013 & $[-0.004876,\,-0.003135]$ & -2.705 & $<0.002$ \\
1K-15 & 1K-15-YaRN & 0.009458 & $[0.002638,\,0.020730]$ & 0.550 & $<0.002$ \\
1K-15 & 10K-15 & -0.005841 & $[-0.007060,\,-0.004815]$ & -3.039 & $<0.002$ \\
1K-40 & 1K-40-YaRN & 0.025028 & $[0.013850,\,0.043445]$ & 0.926 & $<0.002$ \\
1K-40 & 10K-40 & -0.003405 & $[-0.004057,\,-0.002591]$ & -2.749 & $<0.002$ \\
10K-15 & 10K-40 & 0.000379 & $[0.000262,\,0.000496]$ & 1.865 & $<0.002$ \\
\midrule
\multicolumn{6}{l}{\textit{Panel D. Masked-nucleotide Accuracy ($\mathit{Acc}_{\mathrm{mask}}\uparrow$)}} \\
AIDO-CDS & AIDO-CDS-YaRN & 0.042113 & $[0.026675,\,0.056473]$ & 1.686 & $<0.002$ \\
AIDO-CDS-YaRN & 10K-15 & 0.231659 & $[0.149527,\,0.328922]$ & 1.471 & $<0.002$ \\
AIDO-CDS-YaRN & HydraRNA & 0.050845 & $[0.032228,\,0.071879]$ & 1.461 & $<0.002$ \\
HydraRNA & 10K-15 & 0.180814 & $[0.107878,\,0.258662]$ & 1.351 & $<0.002$ \\
RiboGRAM & 10K-15 & 0.234873 & $[0.154680,\,0.329674]$ & 1.532 & $<0.002$ \\
RNAverse & 10K-15 & 0.323938 & $[0.241728,\,0.417506]$ & 2.087 & $<0.002$ \\
1K-15 & 1K-15-YaRN & 0.083339 & $[0.063383,\,0.105623]$ & 2.371 & $<0.002$ \\
1K-15 & 10K-15 & 0.298940 & $[0.216059,\,0.392425]$ & 1.940 & $<0.002$ \\
1K-40 & 1K-40-YaRN & 0.117200 & $[0.091556,\,0.144157]$ & 2.627 & $<0.002$ \\
1K-40 & 10K-40 & 0.319634 & $[0.231481,\,0.423203]$ & 1.968 & $<0.002$ \\
10K-15 & 10K-40 & -0.003118 & $[-0.007930,\,0.000924]$ & -0.420 & 0.149507 \\
\bottomrule
\end{tabular}
\end{table}

\FloatBarrier

\subsection{Attention Restriction}
\label{app:attention-restriction}

For a model with a pretrained context length of $L_{\mathrm{p}}$ tokens, attention from a query position $i$ to a key position $j$ is permitted only when $|i-j|\leq L_{\mathrm{p}}-1$, with boundary tokens counted as positions. The restriction is applied at inference in every attention layer; model weights, position embeddings, and, where used, YaRN scaling are unchanged. $L_{\mathrm{p}}$ is 1,024 for AIDO.RNA-CDS, RiboGRAM, RNAverse, and the \textsc{RiboSpan}-1K variants, 4,096 for HydraRNA~\citep{li2025hydrarna}, and 10,240 for the \textsc{RiboSpan}-10K variants. HydraRNA uses attention in two of its twelve layers; only these are restricted, and its state-space layers are left as they are. The representation endpoints and masked-nucleotide accuracy are recomputed with the same sampled and masked positions as in the unrestricted evaluation. For the span sweep, the \textsc{RiboSpan}-10K variants are additionally evaluated with $L_{\mathrm{p}}$ replaced by 512, 1,024, 2,048, 4,096, and 8,192.

\newpage

Table~\ref{tab:attention-restriction} compares restricted and unrestricted attention at 10,240~nt, where $\Delta$ is the restricted value minus the unrestricted value on the same transcript pairs. Confidence intervals, $d_z$, and BH-adjusted $q$-values follow Appendix~\ref{app:statistical-analysis}, with $\delta_k$ taken as the within-model difference on the $k$th transcript pair and the adjustment applied within each endpoint across all models and length groups. For the \textsc{RiboSpan}-10K variants, the restriction removes only position pairs more than 10,239 tokens apart, and the resulting differences are at the level of numerical noise.

\begin{table}[H]
\centering
\caption{Effect of restricting attention to the pretrained span at 10,240~nt.}
\label{tab:attention-restriction}
\small
\setlength{\tabcolsep}{4pt}
\begin{tabular}{lrrrcrr}
\toprule
Model & Unrestricted & Restricted & $\Delta$ & 95\% CI & $d_z$ & BH $q$ \\
\midrule
\multicolumn{7}{l}{\textit{Panel A. Additional Context Separation ($\Delta\mathit{CS}\uparrow$)}} \\
HydraRNA & 0.313846 & 0.307765 & -0.006081 & $[-0.008131,\,-0.003586]$ & -1.606 & $<0.002$ \\
RiboGRAM & 0.417623 & 0.438528 & 0.020905 & $[0.011767,\,0.028758]$ & 1.436 & $<0.002$ \\
RNAverse & 0.066001 & 0.383584 & 0.317583 & $[0.284995,\,0.353399]$ & 5.168 & $<0.002$ \\
AIDO-CDS & 0.208178 & 0.505050 & 0.296872 & $[0.278247,\,0.314682]$ & 9.716 & $<0.002$ \\
AIDO-CDS-YaRN & 0.446120 & 0.511306 & 0.065185 & $[0.056458,\,0.073079]$ & 4.672 & $<0.002$ \\
1K-15 & 0.221320 & 0.418301 & 0.196981 & $[0.154650,\,0.234757]$ & 2.822 & $<0.002$ \\
1K-15-YaRN & 0.451068 & 0.461103 & 0.010035 & $[-0.010060,\,0.029583]$ & 0.301 & 0.424439 \\
1K-40 & 0.186200 & 0.428535 & 0.242335 & $[0.196352,\,0.284895]$ & 3.122 & $<0.002$ \\
1K-40-YaRN & 0.406227 & 0.451076 & 0.044849 & $[0.025488,\,0.061227]$ & 1.475 & $<0.002$ \\
10K-15 & 0.405962 & 0.405946 & -0.000016 & $[-0.000046,\,0.000003]$ & -0.335 & 0.362557 \\
10K-40 & 0.405777 & 0.405791 & 0.000014 & $[-0.000039,\,0.000072]$ & 0.157 & 0.759089 \\
\midrule
\multicolumn{7}{l}{\textit{Panel B. Cross-region Same-base Similarity ($C_{\mathrm{cross}}\downarrow$)}} \\
HydraRNA & 0.521605 & 0.522310 & 0.000706 & $[-0.002330,\,0.003228]$ & 0.150 & 0.764729 \\
RiboGRAM & 0.404581 & 0.349241 & -0.055340 & $[-0.065182,\,-0.046461]$ & -3.440 & $<0.002$ \\
RNAverse & 0.820777 & 0.424582 & -0.396195 & $[-0.423000,\,-0.374203]$ & -9.659 & $<0.002$ \\
AIDO-CDS & 0.660208 & 0.235812 & -0.424396 & $[-0.442802,\,-0.405333]$ & -13.021 & $<0.002$ \\
AIDO-CDS-YaRN & 0.317370 & 0.252399 & -0.064971 & $[-0.072104,\,-0.058834]$ & -5.700 & $<0.002$ \\
1K-15 & 0.698757 & 0.285413 & -0.413344 & $[-0.426098,\,-0.397303]$ & -16.709 & $<0.002$ \\
1K-15-YaRN & 0.334152 & 0.284899 & -0.049253 & $[-0.068166,\,-0.034768]$ & -1.777 & $<0.002$ \\
1K-40 & 0.744354 & 0.275021 & -0.469333 & $[-0.489565,\,-0.447001]$ & -12.810 & $<0.002$ \\
1K-40-YaRN & 0.361375 & 0.290957 & -0.070418 & $[-0.090368,\,-0.051391]$ & -2.038 & $<0.002$ \\
10K-15 & 0.302668 & 0.302670 & 0.000002 & $[-0.000012,\,0.000018]$ & 0.074 & 0.886028 \\
10K-40 & 0.299277 & 0.299267 & -0.000010 & $[-0.000048,\,0.000036]$ & -0.156 & 0.764729 \\
\midrule
\multicolumn{7}{l}{\textit{Panel C. Relative-distal Diffusion ($D_{\mathrm{distal}}$)}} \\
HydraRNA & 0.000667 & 0.000275 & -0.000392 & $[-0.000610,\,-0.000211]$ & -1.120 & $<0.002$ \\
RiboGRAM & 0.002418 & 0.000024 & -0.002394 & $[-0.002738,\,-0.002064]$ & -4.242 & $<0.002$ \\
RNAverse & 0.004798 & 0.000127 & -0.004671 & $[-0.005524,\,-0.003821]$ & -3.213 & $<0.002$ \\
AIDO-CDS & 0.004666 & 0.000005 & -0.004661 & $[-0.004964,\,-0.004324]$ & -8.498 & $<0.002$ \\
AIDO-CDS-YaRN & 0.025390 & 0.000051 & -0.025338 & $[-0.031690,\,-0.019299]$ & -2.471 & $<0.002$ \\
1K-15 & 0.006626 & 0.000008 & -0.006618 & $[-0.007712,\,-0.005603]$ & -3.623 & $<0.002$ \\
1K-15-YaRN & 0.016084 & 0.000023 & -0.016061 & $[-0.027214,\,-0.009437]$ & -0.948 & $<0.002$ \\
1K-40 & 0.004569 & 0.000009 & -0.004560 & $[-0.005057,\,-0.003983]$ & -4.940 & $<0.002$ \\
1K-40-YaRN & 0.029597 & 0.000058 & -0.029539 & $[-0.048431,\,-0.018157]$ & -1.095 & $<0.002$ \\
10K-15 & 0.000785 & 0.000788 & 0.000002 & $[0.000000,\,0.000006]$ & 0.468 & 0.078755 \\
10K-40 & 0.001164 & 0.001167 & 0.000003 & $[0.000000,\,0.000006]$ & 0.461 & 0.018791 \\
\midrule
\multicolumn{7}{l}{\textit{Panel D. Masked-nucleotide Accuracy ($\mathit{Acc}_{\mathrm{mask}}\uparrow$)}} \\
HydraRNA & 0.427756 & 0.444308 & 0.016552 & $[0.009738,\,0.026216]$ & 1.188 & 0.002453 \\
RiboGRAM & 0.373698 & 0.392078 & 0.018380 & $[0.012062,\,0.025060]$ & 1.646 & 0.002453 \\
RNAverse & 0.284632 & 0.296706 & 0.012073 & $[-0.007485,\,0.028893]$ & 0.373 & 0.398051 \\
AIDO-CDS & 0.334798 & 0.433839 & 0.099041 & $[0.083758,\,0.117551]$ & 3.457 & 0.002453 \\
AIDO-CDS-YaRN & 0.376911 & 0.390475 & 0.013564 & $[0.005320,\,0.021388]$ & 0.993 & 0.004693 \\
1K-15 & 0.309631 & 0.496208 & 0.186577 & $[0.155423,\,0.220801]$ & 3.321 & 0.002453 \\
1K-15-YaRN & 0.392970 & 0.440142 & 0.047172 & $[0.030254,\,0.066980]$ & 1.504 & 0.002453 \\
1K-40 & 0.285818 & 0.490612 & 0.204794 & $[0.171426,\,0.236915]$ & 3.657 & 0.002453 \\
1K-40-YaRN & 0.403019 & 0.424114 & 0.021095 & $[0.011178,\,0.030621]$ & 1.222 & 0.002453 \\
10K-15 & 0.608570 & 0.608115 & -0.000455 & $[-0.000975,\,0.000065]$ & -0.524 & 0.152281 \\
10K-40 & 0.605452 & 0.605389 & -0.000063 & $[-0.000907,\,0.000651]$ & -0.047 & 0.980510 \\
\bottomrule
\end{tabular}
\end{table}

Table~\ref{tab:restricted-paired} compares masked-nucleotide accuracy between models under restriction at 10,240~nt, with $\Delta$ defined as the value for Model B minus that for Model A. Across the four length groups from 2,048 to 10,240~nt, all 72 comparisons between a native 10K variant and another model configuration give $q<0.002$, with differences between 0.066 and 0.436.

\begin{table}[H]
\centering
\caption{Paired comparisons of masked-nucleotide accuracy under attention restriction at 10,240~nt.}
\label{tab:restricted-paired}
\small
\begin{tabular}{llrccr}
\toprule
Model A & Model B & $\Delta$ (B$-$A) & 95\% CI & $d_z$ & BH $q$ \\
\midrule
HydraRNA & 10K-15 & 0.163807 & $[0.094850,\,0.236916]$ & 1.332 & $<0.002$ \\
RiboGRAM & 10K-15 & 0.216037 & $[0.135205,\,0.311415]$ & 1.444 & $<0.002$ \\
RNAverse & 10K-15 & 0.311409 & $[0.223313,\,0.418147]$ & 1.892 & $<0.002$ \\
AIDO-CDS & 10K-15 & 0.174276 & $[0.100048,\,0.260510]$ & 1.270 & $<0.002$ \\
AIDO-CDS-YaRN & 10K-15 & 0.217640 & $[0.137051,\,0.317805]$ & 1.427 & $<0.002$ \\
1K-15 & 10K-15 & 0.111907 & $[0.055913,\,0.176739]$ & 1.074 & $<0.002$ \\
1K-15-YaRN & 10K-15 & 0.167973 & $[0.102490,\,0.241632]$ & 1.436 & $<0.002$ \\
1K-40 & 10K-15 & 0.117503 & $[0.061503,\,0.188555]$ & 1.101 & $<0.002$ \\
1K-40-YaRN & 10K-15 & 0.184001 & $[0.121101,\,0.260928]$ & 1.534 & $<0.002$ \\
10K-40 & 10K-15 & 0.002726 & $[-0.001438,\,0.007088]$ & 0.373 & 0.219665 \\
\bottomrule
\end{tabular}
\end{table}

Table~\ref{tab:span-sweep} reports the span sweep of the \textsc{RiboSpan}-10K variants at 10,240~nt, where $\Delta\mathit{Acc}_{\mathrm{mask}}$ is the accuracy at the given span minus that at the native span. Accuracy decreases significantly for spans of 2,048 tokens or fewer, whereas the decrease at 4,096 tokens is small and not significant. $\Delta\mathit{CS}$ and $C_{\mathrm{cross}}$ change only slightly across spans, and $D_{\mathrm{distal}}$ approaches zero as the span narrows. The shorter length groups show the same ordering.

\begin{table}[H]
\centering
\caption{Attention span sweep of the native 10K \textsc{RiboSpan} variants at 10,240~nt.}
\label{tab:span-sweep}
\small
\setlength{\tabcolsep}{3.5pt}
\begin{tabular}{rrrrrrcrr}
\toprule
Span & $\Delta\mathit{CS}$ & $C_{\mathrm{cross}}$ & $D_{\mathrm{distal}}$ & $\mathit{Acc}_{\mathrm{mask}}$ & $\Delta\mathit{Acc}_{\mathrm{mask}}$ & 95\% CI & $d_z$ & BH $q$ \\
\midrule
\multicolumn{9}{l}{\textit{Panel A. \textsc{RiboSpan}-10K-15}} \\
512 & 0.444179 & 0.297401 & 0.000010 & 0.496014 & -0.112101 & $[-0.183440,\,-0.049117]$ & -0.979 & 0.002570 \\
1,024 & 0.426803 & 0.302667 & 0.000012 & 0.526571 & -0.081544 & $[-0.138946,\,-0.027862]$ & -0.835 & 0.002570 \\
2,048 & 0.415861 & 0.300230 & 0.000088 & 0.560064 & -0.048051 & $[-0.089805,\,-0.011243]$ & -0.709 & 0.002570 \\
4,096 & 0.406774 & 0.301676 & 0.000754 & 0.598985 & -0.009130 & $[-0.020359,\,0.000272]$ & -0.502 & 0.092302 \\
8,192 & 0.405740 & 0.302626 & 0.000941 & 0.608895 & 0.000780 & $[-0.000590,\,0.002150]$ & 0.328 & 0.399800 \\
10,240 & 0.405946 & 0.302670 & 0.000788 & 0.608115 & -- & -- & -- & -- \\
\midrule
\multicolumn{9}{l}{\textit{Panel B. \textsc{RiboSpan}-10K-40}} \\
512 & 0.437311 & 0.299457 & 0.000011 & 0.495570 & -0.109819 & $[-0.182624,\,-0.044255]$ & -0.908 & 0.002570 \\
1,024 & 0.420038 & 0.306799 & 0.000014 & 0.520779 & -0.084609 & $[-0.148429,\,-0.029446]$ & -0.813 & 0.002570 \\
2,048 & 0.411830 & 0.302379 & 0.000141 & 0.557641 & -0.047748 & $[-0.097778,\,-0.010628]$ & -0.649 & 0.003998 \\
4,096 & 0.404155 & 0.301149 & 0.001245 & 0.595205 & -0.010183 & $[-0.024527,\,0.000062]$ & -0.508 & 0.091591 \\
8,192 & 0.405484 & 0.299262 & 0.001312 & 0.605387 & -0.000002 & $[-0.001438,\,0.001557]$ & -0.001 & 1.000000 \\
10,240 & 0.405791 & 0.299267 & 0.001167 & 0.605389 & -- & -- & -- & -- \\
\bottomrule
\end{tabular}
\end{table}

\FloatBarrier

\section{RNA Type Representation Benchmark Details}
\label{app:rna-type-representation}

\subsection{Evaluation Set and Model Inputs}
\label{app:rna-type-inputs}

The representation benchmark uses the 90,000 sequences from the held-out \textsc{RiboSpan} pretraining test set. Sequences are normalized to the A/C/G/T/N alphabet and capped at 10,240~nt before model-specific processing; 417 sequences exceeding this limit are represented by their first 10,240 nucleotides. Models with shorter supported input lengths receive the corresponding prefix, and sequence order is fixed across all models. Table~\ref{tab:rna-type-model-configs} summarizes the model configurations used in the benchmark.

\begin{table}[H]
\centering
\caption{Model configurations used in the RNA-type representation benchmark.}
\label{tab:rna-type-model-configs}
\small
\begin{tabular}{lccc}
\toprule
Model & Layers & Hidden Dim. & Context Length (tokens) \\
\midrule
RNA-FM~\citep{chen2022rnafm} & 12 & 640 & 1,024 \\
RiNALMo~\citep{penic2025rinalmo} & 33 & 1,280 & 1,024 \\
AIDO.RNA-CDS~\citep{zou2024aido} & 32 & 2,048 & 1,024 \\
HydraRNA~\citep{li2025hydrarna} & 12 & 1,024 & 10,240 \\
RiboGRAM~\citep{wang2026ribogram} & 12 & 1,280 & 1,024 \\
RNAverse~\citep{ma2026rnaverse} & 30 & 640 & 1,024 \\
\textsc{RiboSpan}-10K-15 / 10K-40 & 32 & 2,048 & 10,240 \\
\bottomrule
\end{tabular}
\end{table}

\subsection{Frozen Sequence Representations}
\label{app:frozen-representations}

For each sequence, final-layer hidden states over valid nucleotide positions are mean-pooled following the definition in the main text, with boundary and padding tokens excluded. The resulting sequence representations are evaluated in FP32 without feature standardization, PCA, or any learned projection.

\subsection{Evaluation Label Spaces}
\label{app:representation-label-spaces}

The RNA-type evaluations are constructed from the 90,000-sequence held-out test split of the pretraining corpus. Table~\ref{tab:representation-label-composition} summarizes RNA-type composition, Functional mapping, and class sizes in the Long RNA evaluation. Overall Biotype retains the 25 RNA types represented by at least 20 sequences, yielding 89,955 sequences.

The Functional evaluation groups rRNA, tRNA, and tmRNA as housekeeping RNAs; lncRNA, snoRNA, miRNA, pre-miRNA, siRNA, snRNA, and piRNA as regulatory RNAs; and mRNA as coding RNA. These groups contain 20,997, 29,895, and 10,000 sequences, respectively. The Regulatory evaluation uses the seven regulatory RNA types directly as separate labels.

The Long RNA evaluation first selects sequences by original length greater than 1,024~nt and then retains RNA types represented by at least 20 sequences within this subset. This yields 17,130 sequences across antisense RNA, lncRNA, mRNA, miscRNA, ncRNA, rRNA, sRNA, and others.

For the Rfam evaluation, each mapped sequence is assigned to the Rfam family with the lowest E-value, using bit score to resolve ties. The 20 most frequent families are retained, yielding 33,000 sequences. Nearest-neighbor retrieval is performed independently within each label space, with the query sequence itself excluded.

\begin{table}[H]
\centering
\caption{RNA-type composition across Overall, Functional, and Long RNA evaluations.}
\label{tab:representation-label-composition}
\small
\begin{tabular}{lrrl}
\toprule
RNA Type & Test-set $n$ & Long RNA $n$ & Functional Class \\
\midrule
lncRNA              & 10,000 & 4,191 & Regulatory \\
mRNA                & 10,000 & 8,771 & Coding \\
miscRNA             & 10,000 & 1,113 & -- \\
rRNA                & 10,000 & 2,031 & Housekeeping \\
tRNA                & 10,000 & --    & Housekeeping \\
SRP RNA             & 5,000  & --    & -- \\
ncRNA               & 5,000  & 83    & -- \\
piRNA               & 5,000  & --    & Regulatory \\
sRNA                & 5,000  & 834   & -- \\
snRNA               & 5,000  & --    & Regulatory \\
snoRNA              & 5,000  & --    & Regulatory \\
pre-miRNA           & 3,887  & --    & Regulatory \\
hammerhead ribozyme & 1,273  & --    & -- \\
tmRNA               & 997    & --    & Housekeeping \\
RNaseP RNA          & 816    & --    & -- \\
miRNA               & 767    & --    & Regulatory \\
antisense RNA       & 405    & 56    & -- \\
precursor RNA       & 346    & --    & -- \\
ribozyme            & 271    & --    & -- \\
siRNA               & 241    & --    & Regulatory \\
Y RNA               & 129    & --    & -- \\
scaRNA              & 51     & --    & -- \\
vault RNA           & 29     & --    & -- \\
RNaseMRP RNA        & 26     & --    & -- \\
other               & 717    & 51    & -- \\
\bottomrule
\end{tabular}
\end{table}

\subsection{t-SNE Protocol}
\label{app:tsne-protocol}

For each model, the full atlas is fitted on all 90,000 held-out sequences using t-SNE with two output dimensions, perplexity 30, Euclidean distance, and random seed 42, without feature standardization or PCA. Functional and Regulatory panels reuse the corresponding subsets of the full-atlas coordinates, whereas Long RNA and Rfam are fitted independently on their respective 17,130- and 33,000-sequence evaluation sets using the same settings.

\subsection{Functional Class-Level Analysis}
\label{app:functional-results}

The class-level breakdown in Table~\ref{tab:functional-class-purity} shows strong organization across all three functional classes. The \textsc{RiboSpan}-10K checkpoints achieve the highest neighborhood purity for regulatory RNAs, while HydraRNA attains the highest coding-RNA purity; housekeeping purity remains closely matched across the strongest models.

\begin{table}[H]
\centering
\caption{Class-level neighborhood purity in the Functional evaluation. Bold and underlined values in result tables indicate the best and second-best performance, respectively.}
\label{tab:functional-class-purity}
\footnotesize
\setlength{\tabcolsep}{3.5pt}
\begin{tabular}{lccccccccc}
\toprule
Class & $n$ & RNA-FM & RiNALMo & AIDO-CDS & HydraRNA & RiboGRAM & RNAverse & 10K-15 & 10K-40 \\
\midrule
Coding       & 10,000 & 0.870860 & 0.832980 & 0.872060 & \textbf{0.939680} & 0.855300 & 0.801190 & 0.930560 & \underline{0.932130} \\
Housekeeping & 20,997 & 0.988627 & 0.989289 & 0.963104 & 0.987146 & 0.986831 & 0.979321 & \underline{0.989822} & \textbf{0.990008} \\
Regulatory   & 29,895 & 0.928205 & 0.916183 & 0.936789 & 0.964854 & 0.919261 & 0.911169 & \underline{0.972514} & \textbf{0.972574} \\
\bottomrule
\end{tabular}
\end{table}

Most remaining cross-class errors occur between coding and regulatory RNAs. As shown in Table~\ref{tab:functional-confusion}, both \textsc{RiboSpan}-10K checkpoints yield the lowest total coding-regulatory confusion among the models evaluated.

\begin{table}[H]
\centering
\caption{Major confusion counts in the Functional evaluation. Bold and underlined values in result tables indicate the best and second-best performance, respectively.}
\label{tab:functional-confusion}
\small
\begin{tabular}{lccc}
\toprule
Model & Regulatory $\rightarrow$ Coding & Coding $\rightarrow$ Regulatory & Total \\
\midrule
RNA-FM   & 1,770 & 732 & 2,502 \\
RiNALMo  & 2,241 & 880 & 3,121 \\
AIDO-CDS & 1,019 & 871 & 1,890 \\
HydraRNA & 771 & \textbf{352} & 1,123 \\
RiboGRAM & 1,997 & 765 & 2,762 \\
RNAverse & 2,200 & 1,014 & 3,214 \\
10K-15   & \textbf{516} & 427 & \underline{943} \\
10K-40   & \underline{519} & \underline{415} & \textbf{934} \\
\bottomrule
\end{tabular}
\end{table}

\subsection{Long RNA Class-Level Analysis}
\label{app:long-rna-results}

The complete class-level results in Table~\ref{tab:long-rna-class-purity} show that the \textsc{RiboSpan}-10K checkpoints achieve the highest neighborhood purity in six of the eight long-RNA classes, including lncRNA, mRNA, miscRNA, and sRNA.

\begin{table}[H]
\centering
\caption{Class-level neighborhood purity in the Long RNA evaluation. Bold and underlined values in result tables indicate the best and second-best performance, respectively.}
\label{tab:long-rna-class-purity}
\footnotesize
\setlength{\tabcolsep}{3.5pt}
\begin{tabular}{lccccccccc}
\toprule
RNA Type & $n$ & RNA-FM & RiNALMo & AIDO-CDS & HydraRNA & RiboGRAM & RNAverse & 10K-15 & 10K-40 \\
\midrule
Antisense RNA & 56    & 0.033929 & 0.008929 & 0.028571 & 0.039286 & 0.026786 & 0.014286 & \textbf{0.057143} & \underline{0.051786} \\
lncRNA        & 4,191 & 0.622429 & 0.489167 & 0.685898 & 0.763803 & 0.602672 & 0.505082 & \underline{0.782582} & \textbf{0.784586} \\
mRNA          & 8,771 & 0.838627 & 0.802679 & 0.862171 & 0.916133 & 0.830019 & 0.764120 & \underline{0.917649} & \textbf{0.918915} \\
miscRNA       & 1,113 & 0.352022 & 0.336208 & 0.584097 & 0.681491 & 0.386164 & 0.301797 & \textbf{0.727853} & \underline{0.723270} \\
ncRNA         & 83    & \textbf{0.375904} & 0.295181 & 0.271084 & \underline{0.309639} & 0.289157 & 0.237349 & 0.292771 & 0.302410 \\
rRNA          & 2,031 & 0.941310 & \underline{0.949926} & 0.948104 & \textbf{0.966420} & 0.944855 & 0.949237 & 0.947070 & 0.945987 \\
sRNA          & 834   & 0.200600 & 0.141847 & 0.190288 & 0.238729 & 0.172542 & 0.144604 & \textbf{0.272182} & \underline{0.269424} \\
Other         & 51    & 0.013725 & 0.017647 & 0.021569 & 0.050980 & 0.013725 & 0.013725 & \textbf{0.162745} & \underline{0.092157} \\
\bottomrule
\end{tabular}
\end{table}

The confusion counts in Table~\ref{tab:long-rna-confusion} further characterize the major error patterns among long-RNA classes. Both \textsc{RiboSpan}-10K checkpoints yield the lowest total confusion across the four reported directions, while HydraRNA also performs strongly in two individual directions. Together, these results support the value of extended-context RNA modeling, with \textsc{RiboSpan} showing the strongest aggregate separation.

\begin{table}[H]
\centering
\caption{Major confusion counts in the Long RNA evaluation. Bold and underlined values in result tables indicate the best and second-best performance, respectively.}
\label{tab:long-rna-confusion}
\small
\begin{tabular}{lccccc}
\toprule
Model & mRNA $\rightarrow$ lncRNA & lncRNA $\rightarrow$ mRNA
& miscRNA $\rightarrow$ mRNA & miscRNA $\rightarrow$ lncRNA & Total \\
\midrule
RNA-FM   & 619 & 766   & 619 & 129 & 2,133 \\
RiNALMo  & 784 & 1,495 & 630 & 124 & 3,033 \\
AIDO-CDS & 686 & 452   & 264 & 156 & 1,558 \\
HydraRNA & \textbf{289} & 291 & 243 & \textbf{72} & 895 \\
RiboGRAM & 676 & 880 & 565 & 124 & 2,245 \\
RNAverse & 963 & 1,346 & 641 & 153 & 3,103 \\
10K-15   & 345 & \underline{210} & \textbf{199} & 83 & \underline{837} \\
10K-40   & \underline{330} & \textbf{203} & \underline{204} & \underline{79} & \textbf{816} \\
\bottomrule
\end{tabular}
\end{table}

\section{RNAGym Evaluation Details}
\label{app:rnagym}

We evaluate zero-shot mutation fitness using the processed deep-mutational-scanning assays distributed with RNAGym~\citep{arora2025rnagym}. After normalizing U to T and retaining parsable substitution variants, the evaluation contains 70 assays and 1,117,295 variants spanning five RNA categories: aptamer, ribozyme, tRNA, mRNA-splicing, and mRNA-coding (Table~\ref{tab:rnagym-composition}).

\begin{table}[H]
\centering
\caption{Composition of the RNAGym mutation-fitness evaluation.}
\label{tab:rnagym-composition}
\small
\begin{tabular}{llrrrrr}
\toprule
RNA Type & Description & Assays & Mutants & Single & Multiple & WT Length (nt) \\
\midrule
Aptamer & Target binding ability & 2 & 3,069 & 446 & 2,623 & 70--82 \\
Ribozyme & Cleavage and splicing & 26 & 758,333 & 4,118 & 754,215 & 45--425 \\
Transfer RNA (tRNA) & Stability and growth & 3 & 95,202 & 434 & 94,768 & 72--105 \\
Messenger RNA (mRNA) & Splicing ability & 2 & 5,722 & 330 & 5,392 & 51--63 \\
Messenger RNA (mRNA) & Coding mRNA fitness & 37 & 254,969 & 40,468 & 214,501 & 144--5,592 \\
\midrule
Total &  & 70 & 1,117,295 & 45,796 & 1,071,499 & 45--5,592 \\
\bottomrule
\end{tabular}
\end{table}

\paragraph{Masked-marginal scoring.}
All evaluated bidirectional masked language models use the same wild-type-background scoring procedure. For a mutation set with positions $M$, all mutated positions are masked simultaneously while all remaining positions retain the wild-type sequence. The mutation score is

\begin{equation}
s(x^{\mathrm{mut}})
=
\sum_{i\in M}
\left[
\log p_{\theta}
\left(
x_i^{\mathrm{mut}}
\mid
\widetilde{x}
\right)
-
\log p_{\theta}
\left(
x_i^{\mathrm{WT}}
\mid
\widetilde{x}
\right)
\right].
\label{eq:rnagym-masked-marginal}
\end{equation}

This yields an order-independent masked-marginal score for multiple substitutions under the wild-type background. Sequences exceeding a model's native context are evaluated using a mutation-centered window containing all mutated positions; otherwise, the full sequence is used directly.

\paragraph{Metrics and aggregation.}
Metric definitions and score aggregation follow the official RNAGym evaluation procedure exactly. For each assay, the reported Spearman $r$ score is the absolute Spearman correlation between experimental fitness $y$ and model score $s$,

\begin{equation}
\rho
=
\left|
\operatorname{Spearman}(y,s)
\right|.
\end{equation}

AUROC is computed after binarizing experimental fitness at the assay-specific median and is folded with respect to score direction,

\begin{equation}
\operatorname{AUROC}
=
\max
\left(
\operatorname{AUC},
1-\operatorname{AUC}
\right).
\end{equation}

MCC is computed after binarizing experimental fitness and model scores at their respective medians,

\begin{equation}
\operatorname{MCC}
=
\left|
\operatorname{MCC}
\left(
y>\operatorname{median}(y),
s>\operatorname{median}(s)
\right)
\right|.
\end{equation}

Following the RNAGym evaluation procedure, each metric is first computed independently for every assay, then averaged equally within each RNA category, and finally averaged equally across the five categories. This category-balanced aggregation prevents categories containing many more assays from dominating the overall score.

\begin{table}[H]
\centering
\caption{RNAGym performance by RNA category. Bold and underlined values indicate the best and second-best performance among the evaluated encoder-only models, respectively.}
\label{tab:rnagym-by-type}
\footnotesize
\begin{tabularx}{0.9\textwidth}{ll*{5}{>{\centering\arraybackslash}X}}
\toprule
\multirow{2}{*}{Model} 
& \multirow{2}{*}{Metric} 
& \multicolumn{2}{c}{mRNA} 
& \multirow{2}{*}{tRNA} 
& \multirow{2}{*}{Aptamer} 
& \multirow{2}{*}{Ribozyme} \\
\cmidrule(lr){3-4}
& & Splicing & Fitness & & & \\
\midrule

\multirow{3}{*}{RNA-FM}
& Spearman $r$ & 0.2488 & 0.0582 & 0.3857 & 0.0150 & 0.1292 \\
& AUROC        & 0.6320 & 0.5331 & 0.6920 & 0.5156 & 0.5646 \\
& MCC          & 0.1963 & 0.0479 & 0.2902 & 0.0072 & 0.0982 \\
\midrule

\multirow{3}{*}{RiNALMo}
& Spearman $r$ & 0.3557 & 0.0630 & 0.4581 & 0.0463 & 0.1167 \\
& AUROC        & 0.6881 & 0.5312 & 0.7247 & 0.5231 & 0.5582 \\
& MCC          & 0.2626 & 0.0457 & 0.3386 & 0.0245 & 0.0846 \\
\midrule

\multirow{3}{*}{AIDO-CDS}
& Spearman $r$ & 0.2415 & 0.1472 & 0.4163 & 0.0642 & 0.1295 \\
& AUROC        & 0.6371 & 0.5758 & 0.7036 & 0.5270 & 0.5686 \\
& MCC          & 0.2415 & 0.1094 & 0.3130 & 0.0513 & 0.0939 \\
\midrule

\multirow{3}{*}{HydraRNA}
& Spearman $r$ & 0.3165 & 0.1122 & 0.4148 & 0.0298 & 0.0872 \\
& AUROC        & 0.6666 & 0.5599 & 0.7021 & 0.5231 & 0.5429 \\
& MCC          & 0.2732 & 0.0881 & 0.3081 & 0.0218 & 0.0625 \\
\midrule

\multirow{3}{*}{RiboGRAM}
& Spearman $r$ & 0.2864 & 0.0678 & 0.4448 & 0.0270 & 0.1529 \\
& AUROC        & 0.6589 & 0.5376 & 0.7195 & 0.5131 & 0.5801 \\
& MCC          & 0.2528 & 0.0544 & 0.3311 & 0.0144 & 0.1193 \\
\midrule

\multirow{3}{*}{RNAverse}
& Spearman $r$ & \textbf{0.4081} & 0.1218 & 0.4014 & 0.0257 & 0.1090 \\
& AUROC        & \underline{0.7019} & 0.5635 & 0.6998 & 0.5187 & 0.5546 \\
& MCC          & \underline{0.3182} & 0.0907 & 0.2988 & 0.0168 & 0.0797 \\
\midrule

\multirow{3}{*}{LIMBER-macro}
& Spearman $r$ & 0.3055 & 0.0623 & 0.4541 & 0.0261 & \textbf{0.2044} \\
& AUROC        & 0.6687 & 0.5340 & 0.7249 & 0.5293 & \textbf{0.6008} \\
& MCC          & 0.2349 & 0.0481 & 0.3368 & 0.0511 & \textbf{0.1492} \\
\midrule

\multirow{3}{*}{1K-15}
& Spearman $r$ & 0.3437 & 0.1486 & \textbf{0.4781} & 0.0589 & 0.1753 \\
& AUROC        & 0.6813 & \underline{0.5767} & \textbf{0.7367} & 0.5331 & 0.5847 \\
& MCC          & 0.2734 & \textbf{0.1127} & \textbf{0.3508} & 0.0497 & \underline{0.1274} \\
\midrule

\multirow{3}{*}{1K-40}
& Spearman $r$ & \underline{0.4059} & 0.1346 & 0.4639 & \underline{0.0794} & \underline{0.1781} \\
& AUROC        & \textbf{0.7102} & 0.5706 & 0.7305 & \underline{0.5421} & \underline{0.5865} \\
& MCC          & \textbf{0.3372} & 0.1027 & 0.3471 & 0.0462 & 0.1189 \\
\midrule

\multirow{3}{*}{10K-15}
& Spearman $r$ & 0.2913 & \underline{0.1499} & 0.4589 & 0.0600 & 0.1541 \\
& AUROC        & 0.6666 & 0.5762 & 0.7266 & 0.5391 & 0.5759 \\
& MCC          & 0.2563 & 0.1106 & 0.3392 & \underline{0.0661} & 0.1108 \\
\midrule

\multirow{3}{*}{10K-40}
& Spearman $r$ & 0.2939 & \textbf{0.1521} & \underline{0.4694} & \textbf{0.0818} & 0.1535 \\
& AUROC        & 0.6559 & \textbf{0.5784} & \underline{0.7317} & \textbf{0.5471} & 0.5770 \\
& MCC          & 0.2399 & \underline{0.1114} & \underline{0.3482} & \textbf{0.0941} & 0.1163 \\
\bottomrule
\end{tabularx}
\end{table}

Table~\ref{tab:rnagym-by-type} shows that the \textsc{RiboSpan} checkpoints lead in most categories and exhibit distinct strengths across them, highlighting complementary specialization within the checkpoint family. Among the external models, LIMBER-macro is the strongest on ribozymes, and RNAverse attains the highest Spearman correlation on mRNA splicing.

\section{mRNABench Evaluation Details}
\label{app:mrnabench}

Our evaluation follows the mRNABench linear-probing protocol~\citep{shi2025mrnabench}, including its task definitions, data-splitting strategy, frozen-embedding evaluation, linear probes, and evaluation metrics. HL, MRL, eCLIP, GO, and VEP are obtained and processed through mRNABench v1.2.2, with TE included from the corresponding Morris Lab catalogue entries. The evaluated tasks are summarized in Table~\ref{tab:mrnabench-composition}.

\begin{table}[H]
\centering
\caption{Task composition of the mRNABench full-transcript evaluation.}
\label{tab:mrnabench-composition}
\small
\begin{tabular}{llcccc}
\toprule
Family & Task Type & Unique Subtasks & Mean Length (nt) & Max Length (nt) & Metric \\
\midrule
HL    & Regression             & 2  & 3,549 & 12,288  & Pearson $r$ \\
MRL   & Regression             & 1  & 2,620 & 12,275  & Pearson $r$ \\
TE    & Regression             & 2  & 3,723 & 17,497  & Pearson $r$ \\
eCLIP & Binary classification  & 40 & 3,370 & 205,012 & AUPRC \\
GO    & Multilabel classification & 3 & 3,522 & 12,258 & micro-AUPRC \\
VEP   & Binary classification  & 2  & 3,659 & 28,227  & AUPRC \\
\bottomrule
\end{tabular}
\end{table}

\paragraph{Full-transcript representations.}
Each input uses the complete mature-mRNA sequence. Final-layer nucleotide representations are mean-pooled over all valid positions. Sequences exceeding the model's native context are divided into non-overlapping chunks, which are encoded independently and combined by a length-weighted average of the chunk-level representations. This ensures that every nucleotide contributes equally to the final transcript representation. Most inputs require chunking for the 1K models, whereas only a small fraction do so for the 10K models. Final pooled embeddings are stored in FP32.

\paragraph{Data splits and probes.}
Processed sequence, label, and split tables are fixed before evaluation. Evaluation uses ten fixed splits with seeds [2541, 413, 411, 412, 2547, 321, 421, 311, 2516, 2515], each containing approximately 70\% training, 15\% validation, and 15\% test examples. HL, MRL, eCLIP, GO, and TE use the homology-aware splitting procedure from mRNABench, which keeps mapped NCBI homology groups within the same split, whereas VEP uses the catalogue default random split.

All pretrained model parameters remain frozen. A separate probe is trained for every model, dataset, target, and split. Regression tasks use Ridge regression with cross-validated regularization over $\alpha \in \{10^{-3},10^{-2},10^{-1},1,10\}$; binary classification uses L2-regularized logistic regression; and multilabel classification uses independent logistic-regression classifiers for each label. No feature standardization, PCA, nonlinear projection, or backbone fine-tuning is applied. Following mRNABench, the reported validation metrics are Pearson $r$ for HL, MRL, and TE, AUPRC for eCLIP and VEP, and micro-AUPRC for GO, which consists of multilabel prediction tasks.

\section{RNA Secondary Structure Prediction Details}
\label{app:secondary-structure}

\paragraph{structRFM protocol.}
Following structRFM~\citep{zhu2026structrfm}, a MixedFold prediction head with Zuker decoding is trained on bpRNA TR0 (10,814 sequences) and evaluated on TS0 (1,305 sequences), and trained on RNAStrAlign (37,149 sequences) and evaluated on ArchiveII. Sequences are truncated to 511 nucleotides. For \textsc{RiboSpan}, the backbone is frozen and only the head is trained for 60 epochs, and the epoch is selected as in the structRFM protocol. structRFM is evaluated with its published checkpoints, which are fine-tuned end to end. The \textsc{RiboSpan} heads are evaluated on the ArchiveII sequences of at most 599 nucleotides (3,911 sequences), and all models are compared on these sequences. We report F1 and the interaction network fidelity, $\mathrm{INF}=\sqrt{\mathrm{precision}\times\mathrm{recall}}$, averaged over sequences (Table~\ref{tab:structrfm-protocol}). The comparison does not depend on the selected epoch: for both \textsc{RiboSpan}-10K variants, every one of the last 48 of the 60 epochs exceeds structRFM in mean F1 on ArchiveII, and every one of the last 55 does so on bpRNA TS0.

\begin{table}[H]
\centering
\caption{Secondary structure prediction under the structRFM protocol. Bold and underlined values indicate the best and second-best performance, respectively.}
\label{tab:structrfm-protocol}
\small
\begin{tabular}{llcccc}
\toprule
Test Set & Model & F1 & INF & Precision & Recall \\
\midrule
\multirow{3}{*}{ArchiveII ($n=3{,}911$)}
& structRFM~\citep{zhu2026structrfm} & 0.8877 & 0.8898 & 0.9041 & 0.8800 \\
& \textsc{RiboSpan}-10K-15 & \textbf{0.9098} & \textbf{0.9126} & \textbf{0.9456} & \textbf{0.8855} \\
& \textsc{RiboSpan}-10K-40 & \underline{0.9061} & \underline{0.9088} & \underline{0.9416} & \underline{0.8818} \\
\midrule
\multirow{3}{*}{bpRNA TS0 ($n=1{,}305$)}
& structRFM~\citep{zhu2026structrfm} & 0.6626 & 0.6686 & 0.6346 & 0.7158 \\
& \textsc{RiboSpan}-10K-15 & \underline{0.6935} & \underline{0.6994} & \textbf{0.6650} & \underline{0.7464} \\
& \textsc{RiboSpan}-10K-40 & \textbf{0.6985} & \textbf{0.7049} & \underline{0.6635} & \textbf{0.7603} \\
\bottomrule
\end{tabular}
\end{table}

\paragraph{LIMBER protocol.}
Following the RiNALMo fine-tuning protocol~\citep{penic2025rinalmo} as used by LIMBER~\citep{wang2026limber}, a residual convolutional head predicts base-pairing probabilities from pairwise combinations of nucleotide representations and is trained with binary cross-entropy for 26 epochs, while the backbone is gradually unfrozen. Structures are decoded greedily over canonical base pairs with a probability threshold tuned on the validation set, and performance is measured by F1 with a tolerance of one nucleotide (flexible F1). Sequences longer than 512 nucleotides are excluded. bpRNA uses the TR0, VL0, and TS0 splits (10,814, 1,300, and 1,305 sequences); Rivas uses TrainSetA with 10\% held out for validation (2,649 and 294 sequences) and TestSetB (430 sequences); ArchiveII uses nine leave-one-family-out splits with 3,865 test sequences in total. The checkpoint with the highest validation F1 is evaluated once on the test set, and each model is trained once. Whereas RNAStrAlign and ArchiveII share RNA families in the structRFM protocol, here the test family is absent from training.

Table~\ref{tab:limber-protocol} summarizes the three evaluations, with ArchiveII reported as the mean F1 pooled over the test sequences of all held-out families, and Table~\ref{tab:limber-archiveii-family} lists each family. \textsc{RiboSpan}-10K-15 and 10K-40 exceed LIMBER-macro in seven and six of the nine held-out families, respectively.

\begin{table}[H]
\centering
\caption{Secondary structure prediction under the LIMBER protocol (flexible F1). Bold and underlined values indicate the best and second-best performance, respectively.}
\label{tab:limber-protocol}
\small
\begin{tabular}{lccccc}
\toprule
& \multicolumn{2}{c}{bpRNA TS0} & \multicolumn{2}{c}{Rivas TestSetB} & ArchiveII \\
\cmidrule(lr){2-3}\cmidrule(lr){4-5}\cmidrule(lr){6-6}
Model & Mean & Median & Mean & Median & Pooled Mean \\
\midrule
LIMBER-macro~\citep{wang2026limber} & \textbf{0.7927} & \textbf{0.8889} & \textbf{0.7559} & \textbf{0.8108} & \underline{0.8250} \\
\textsc{RiboSpan}-10K-15 & \underline{0.7798} & \underline{0.8725} & 0.7204 & 0.7670 & \textbf{0.8277} \\
\textsc{RiboSpan}-10K-40 & 0.7741 & 0.8697 & \underline{0.7344} & \underline{0.7853} & 0.8232 \\
\bottomrule
\end{tabular}
\end{table}

\begin{table}[H]
\centering
\caption{Mean flexible F1 by held-out family in the ArchiveII leave-one-family-out evaluation. Bold and underlined values indicate the best and second-best performance, respectively.}
\label{tab:limber-archiveii-family}
\small
\begin{tabular}{lrccc}
\toprule
Family & $n$ & LIMBER-macro & 10K-15 & 10K-40 \\
\midrule
5S         & 1,283 & \textbf{0.8817} & 0.8686 & \underline{0.8703} \\
16S        & 67    & 0.7620 & \underline{0.7846} & \textbf{0.7969} \\
23S        & 15    & 0.8761 & \textbf{0.9023} & \underline{0.8840} \\
tRNA       & 557   & 0.9318 & \textbf{0.9569} & \underline{0.9433} \\
tmRNA      & 462   & \underline{0.8478} & \textbf{0.8551} & 0.8405 \\
SRP        & 918   & 0.7154 & \textbf{0.7430} & \underline{0.7367} \\
RNase P    & 454   & \textbf{0.8304} & 0.7782 & \underline{0.7788} \\
Group I    & 74    & 0.6476 & \underline{0.6957} & \textbf{0.6959} \\
Telomerase & 35    & 0.0243 & \textbf{0.1034} & \underline{0.0957} \\
\bottomrule
\end{tabular}
\end{table}

\paragraph{Pseudoknots.}
We also score crossing base pairs alone, using the same leave-one-family-out models, in the six ArchiveII families whose reference structures contain pseudoknots (5S rRNA, tRNA, and SRP contain none). A base pair is crossing if it is interleaved with another pair of the same structure; following IPknot~\citep{sato2022ipknot}, a predicted crossing pair is correct only if the reference contains exactly the same pair and it is crossing there as well. F1 is computed from the counts pooled over the test sequences of each family, and IPknot v1.0.0, run with default settings on the same sequences, serves as an external reference (Table~\ref{tab:limber-pseudoknot}).

\begin{table}[H]
\centering
\caption{F1 over crossing base pairs in the ArchiveII leave-one-family-out evaluation. $n_{\mathrm{pk}}/n$ is the number of test sequences with a pseudoknotted reference out of all test sequences. Bold and underlined values indicate the best and second-best performance, respectively.}
\label{tab:limber-pseudoknot}
\small
\begin{tabular}{lrrcccc}
\toprule
Family & $n_{\mathrm{pk}}/n$ & Crossing Pairs & IPknot & LIMBER-macro & 10K-15 & 10K-40 \\
\midrule
16S        & 20/67   & 198    & 0.0348 & \underline{0.0736} & \textbf{0.1119} & 0.0711 \\
23S        & 3/15    & 17     & 0.0000 & 0.0075 & \textbf{0.0610} & \underline{0.0541} \\
tmRNA      & 459/462 & 26,153 & 0.2384 & \underline{0.8079} & \textbf{0.8264} & 0.8077 \\
RNase P    & 417/454 & 10,133 & 0.1512 & \textbf{0.7868} & \underline{0.6739} & 0.5229 \\
Group I    & 74/74   & 907    & 0.1358 & \underline{0.4483} & \textbf{0.5108} & 0.3709 \\
Telomerase & 35/35   & 969    & \underline{0.0450} & 0.0176 & \textbf{0.0876} & 0.0190 \\
\bottomrule
\end{tabular}
\end{table}

\FloatBarrier
\clearpage

\pagestyle{plain}

\bibliographystyle{acl_natbib}
\bibliography{bib}

\end{document}